\documentclass[10pt,twocolumn,letterpaper]{article}

\usepackage{wacv}              % To produce the CAMERA-READY version
\usepackage{graphicx}
\usepackage{amsmath}
\usepackage{amssymb}
\usepackage{booktabs}
\usepackage{bbm}
\usepackage{xcolor}
\usepackage{algorithm}
\usepackage{algpseudocode}
\usepackage[title,page]{appendix} % Add 'page' option here
\usepackage[accsupp]{axessibility}  % Improves PDF readability for those with disabilities.

\usepackage[pagebackref,breaklinks,colorlinks]{hyperref}

\usepackage[capitalize]{cleveref}
\crefname{section}{Sec.}{Secs.}
\Crefname{section}{Section}{Sections}
\Crefname{table}{Table}{Tables}
\crefname{table}{Tab.}{Tabs.}

\def\wacvPaperID{650} % *** Enter the WACV Paper ID here
\def\confName{WACV}
\def\confYear{2024}

\newif\ifdraft
\drafttrue

\definecolor{orange}{rgb}{1,0.5,0}
\ifdraft
 \newcommand{\RS}[1]{{\color{red}{\bf RS: #1}}}
 
 \newcommand{\PMN}[1]{{\color{orange}{\bf PMN: #1}}}

 \else
 \newcommand{\RS}[1]{{\color{red}{}}}
 
 \newcommand{\PMN}[1]{{\color{red}{}}}

\fi

\newcommand{\x}{\mathbf{x}}
\newcommand{\z}{\mathbf{z}}
\newcommand{\y}{\mathbf{y}}
\renewcommand{\a}{\mathbf{a}}
\renewcommand{\k}{\mathbf{k}}

\newcommand{\A}{\mathcal{A}}
\newcommand{\U}{\mathcal{U}}
\newcommand{\X}{\mathbf{X}}
\newcommand{\cX}{\mathcal{X}}
\newcommand{\cY}{\mathcal{Y}}
\newcommand{\cZ}{\mathcal{Z}}

\DeclareMathOperator{\simil}{sim}

\newcommand{\COALSamp}{{\emph{COWAL}}}
\newcommand{\COALCenterSamp}{{\emph{COWAL center}}}
\newcommand{\RSamp}{{\emph{Random}}}
\newcommand{\ESamp}{{\emph{Entropy}}}
\newcommand{\TempSamp}{{\emph{Temporal Coverage}}}

\newcommand{\CoreSamp}{{\emph{CoreSet}}}
\newcommand{\CoreEntSamp}{{\emph{CoreSet x Entropy}}}
\newcommand{\MCSamp}{{\emph{MC Dropout}}}
\newcommand{\BALDSamp}{{\emph{BALD}}}
\newcommand{\SASamp}{{\emph{Suggestive Annotation}}}
\newcommand{\VAALSamp}{{\emph{VAAL}}}

\begin{document}

\title{Correlation-aware active learning for surgery video segmentation}

\author{Fei Wu\\
AIMI, University of Bern\\
{\tt\small fei.wu@unibe.ch}
\and
Pablo Marquez-Neila\\
AIMI, University of Bern\\
{\tt\small pablo.marquez@unibe.ch}
\and
Mingyi Zheng\\
Auris Health. Inc, JNJ\\
{\tt\small immingyizheng@gmail.com}
\and
Hedyeh Rafii-Tari\\
Auris Health. Inc, JNJ\\
{\tt\small hrafiita@its.jnj.com}
\and
Raphael Sznitman\\
AIMI, University of Bern\\
{\tt\small raphael.sznitman@unibe.ch}
}

\maketitle

%%%%%%%%% ABSTRACT
\begin{abstract}
Semantic segmentation is a complex task that relies heavily on large amounts of annotated image data. However, annotating such data can be time-consuming and resource-intensive, especially in the medical domain. Active Learning (AL) is a popular approach that can help to reduce this burden by iteratively selecting images for annotation to improve the model performance. In the case of video data, it is important to consider the model uncertainty and the temporal nature of the sequences when selecting images for annotation. This work proposes a novel AL strategy for surgery video segmentation, \COALSamp{}, COrrelation-aWare Active Learning. Our approach involves projecting images into a latent space that has been fine-tuned using contrastive learning and then selecting a fixed number of representative images from local clusters of video frames. We demonstrate the effectiveness of this approach on two video datasets of surgical instruments and three real-world video datasets. The datasets and code will be made publicly available upon receiving necessary approvals.
\end{abstract}

% \begin{figure}[t]
%     \centering
%     \includegraphics[width=1\linewidth]{figures/baseline_comparison_intuitive_v2.pdf}
%    \caption{An overview of our proposed sampling method for video datasets. Each point represents a video frame. Red points are the currently labeled frames, grey points are the unlabeled frames, and green points are the frame to be labeled next. We use K-Means to create image clusters in an embedding space. For each red point, we force a cluster construction to be centered around it. The rest of the clusters would be centered around unlabeled frames. We then query the points closest to the center of green clusters. This approach allows unlabeled frames similar to currently labeled frames to be grouped and lets frames visually different from currently labeled frames be queried.}
% \label{method_overview}
% \end{figure}

% \begin{figure*}[t]
%     \includegraphics[width=1\linewidth]{figures/method_overview.pdf}
%    \caption{Each row shows a video from a dataset we worked on. From left to right is the chronological order. The first and last images of the video segment also display the binary segmentation mask. The second row displays images of our internal dataset and they correspond to views from inside the lung with a surgery tool in it.}
% \label{all_dataset}
% \end{figure*}

%%%%%%%%% BODY TEXT
\section{Introduction}

Minimally invasive surgical robotic systems have seen widespread adoption for surgical procedures. In this context, endoscopic video feeds allow new possibilities to enhance, augment and even partially automate specific tasks. This includes clinical decision support~\cite{gastroscopic_images}, case review~\cite{digestive_organ_recognition}, or even intra-operational fusion~\cite{bodenstedt2018comparative}. However, a key necessity for these applications is the ability to segment surgical instruments~\cite{Islam,Shvets,endovis}. To this end, promising performances for binary segmentation have been shown, but these methods suffer from the need for large amounts of manually annotated data to train subsequent models.

To alleviate the segmentation annotation burden, Active Learning (AL) is a well-established learning strategy~\cite{medical_AL_survey,disparency_AL,warm_start_AL,suggestive_ann,diminishing_uncertainty,DSAL}. Using a large set of unlabeled data, AL iteratively optimizes which data points should be annotated by an oracle (\ie,~expert) to improve the model performance most efficiently. To date, it has been widely applied to different clinical applications such as echocardiography view classification ~\cite{echocardiogram_AL}, diabetic retinopathy detection ~\cite{medAL}, or~surgical phases recognition~\cite{Shi2022}.

Video sequences pose unique challenges for AL methods due to the high correlation between video frames. An example of this can be seen in Fig.~\ref{fig:auris} (rows~2,~4,~5,~6 and~7), where consecutive frames exhibit high visual similarity. Annotating more than one of these redundant frames would bring just a marginal information gain to the segmentation model and, considering the high cost of segmentation annotations, should be avoided by an effective AL~method. However, while typical segmentation AL~methods such as entropy sampling incorporate some implicit or explicit mechanism to prevent redundant samples from being annotated, they are designed under the assumption that image samples are independent and identically distributed, which is not satisfied in the case of video sequences and necessarily leads to suboptimal sample choices. This is especially problematic in deep learning AL methods, where multiple samples sharing similar properties are simultaneously selected at each step. Therefore, video AL methods must be designed to account for the highly correlated nature of video sequences explicitly.

Even though many works have applied AL for videos~\cite{bubblenet,video_caption_AL,video_tracking_AL,video_tracking_AL_2,temporal_coherence_AL,combining_self_training} and semantic segmentation~\cite{suggestive_ann,vae_AL,DSAL,BRATS_AL,viewAL, towards_fewer_ann}, only a few have studied AL methods for semantic segmentation on video datasets~\cite{synthetic_image_AL, best_practices_AL}. Peng~\etal~\cite{synthetic_image_AL} used the EndoVis 2017 dataset~\cite{endovis_2017_dataset}, which contains surgery videos, but they curated the dataset to remove similar frames, hence the challenge within video datasets, and used a sampling strategy based on model uncertainty. Mittai \etal~\cite{best_practices_AL} compared existing AL methods for video datasets. They pointed out the inefficiency of uncertainty-based methods on videos because these methods do not consider the information redundancy in video datasets. They showed that a diversity-based method Core-Set \cite{core_set}, is a better choice of sampling strategy for video datasets. We validate their finding and further show that our approach, combining uncertainty and diversity, can yield better performance for surgery videos. 
 
We propose \COALSamp{}, an AL strategy for video segmentation. Given that it is computationally advantageous to sample multiple images when iteratively training NNs, we hypothesize that the selection of images should consider both the model uncertainty and the temporal nature of video sequences. We propose to project images in a latent space, fine-tuned by contrastive learning, and then select a fixed number of images at each iteration representative of local clusters of video frames. We show experimentally that our approach is superior to standard AL selection strategies.

\section{Related Works}

We present here some of the relevant related works. These consist of AL methods for video applications, methods for semantic segmentation, and general AL methods. 

\subsection{Semantic Segmentation}
Semantic segmentation is the task of classifying each pixel of an image into a defined category. Several AL methods for semantic segmentation have been proposed in the past. Broadly, these can be categorized as methods that sample images ~\cite{suggestive_ann,vae_AL,DSAL,BRATS_AL,synthetic_image_AL} or sample image regions~\cite{towards_fewer_ann, revisting_superpixel,DIAL,CEREALS,MetaBox,RL_AL, self_consistency,viewAL,active_semi_supervised_learning,MEAL, REDAL}. While sampling image regions allows for finer-grained queries (\ie labeling of a group of adjacent pixels in a frame), it often comes at the expense of more complex AL methods with higher performance variance. Instead, this work focuses on image-level sampling, allowing for more scalable AL, and additional annotation costs are marginal in time and effort compared to neural network training time.

% Moreover, an image-level sampling approach ensures that our method is task agnostic, and even though we only experimented on semantic segmentation tasks due to the high annotation cost of this task, our method can well be applied to other tasks. \\
In the category of methods that sample images, Yang \etal \cite{suggestive_ann} first considers a subset $S$ of samples from the unlabeled pool with the highest entropy, usually two times the number of the query budget. They proceed to iteratively select samples from this subset that best cover the unlabeled data distribution in an embedding space. Unlike them, we also take into account the embedding of labeled data and group all embeddings into distinct clusters before sampling the image with the highest uncertainty from each cluster.

Sinha \etal \cite{vae_AL} use adversarial training between a VAE \cite{vae} and a discriminator to generate image embedding using the VAE and having the discriminator learn the difference between embeddings of labeled images and unlabeled images at every AL step. Once the discriminator is trained, they select unlabeled samples that are most different from labeled samples according to the discriminator. Instead of training a VAE at every AL step, we train an embedding model once using contrastive learning to learn image similarity. We then cluster unlabeled images with visually similar labeled images and sample unlabeled images that are not grouped with labeled images.

\subsection{Video Datasets}
Due to the vast amounts of data available in video content, AL learning for video has drawn research interests in recent years~\cite{best_practices_AL,bubblenet,video_caption_AL,video_tracking_AL,video_tracking_AL_2, synthetic_image_AL,temporal_coherence_AL,combining_self_training}. These methods have focused on a wide range of tasks, such as video caption learning \cite{video_caption_AL}, surgical phase classification \cite{LRTD}, video object detection \cite{temporal_coherence_AL}, video tracking \cite{video_tracking_AL,video_tracking_AL_2}, video object segmentation \cite{bubblenet}, and video semantic segmentation \cite{best_practices_AL}.

While Griffin and Corso \cite{bubblenet} developed a method for video object segmentation, this task only requires the annotator to label one frame per video, and a model will learn to track the segmented object of this one frame throughout the video. In contrast, our method handles semantic segmentation and requires the sampling of several frames. Thus the method developed by Griffin and Corso \cite{bubblenet} can not be directly applied in our case.

The work of Mittal \etal~\cite{best_practices_AL} is, however, relevant to ours. Similar to our findings, they have pointed out the inefficiency of the uncertainty-based sampling strategy for datasets with redundant images. They applied AL methods to the A2D2 \cite{A2D2} and Cityscapes \cite{cityscapes} datasets. Even though Cityscapes \cite{cityscapes} comes from video sequences, it is more akin to image datasets because of the curation it went through to remove redundant frames. We thus focus on their result for the A2D2 dataset \cite{A2D2}, and their finding is that a diversity-based sampling strategy Core-Set \cite{core_set} performs better than uncertainty-based methods, which aligns with our experiments. They have nevertheless not tested sampling methods that combine both uncertainty and diversity. We show in this paper a similar study focusing on surgery videos. However, unlike Mittal \etal \cite{best_practices_AL}, who compared existing AL methods for video segmentation, we also propose our own sampling strategy based on uncertainty-diversity.

\begin{figure*}[t]
    \includegraphics[width=1\linewidth]{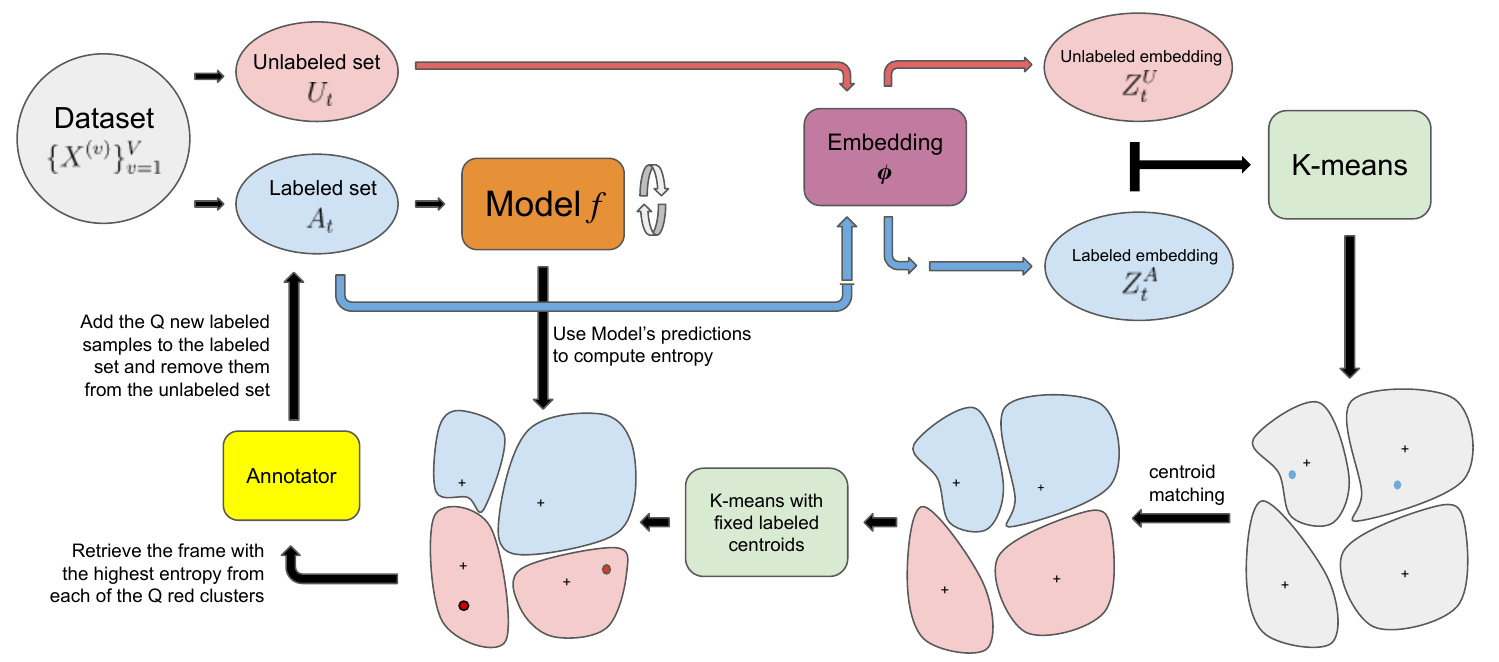}
   \caption{An overview of our approach. A subset of the dataset is labeled to train a model. Once the model has converged, an embedding function projects all the images in a latent space. K-means is then used to create clusters of similar images. We assign all labeled frames to the closest centroids and fix the centroids to those. K-means is applied again over the remaining centroids. Our strategy selects the samples with the highest model entropy per cluster for annotation.}
\label{fig:results_auris}
\end{figure*}

\subsection{Active Learning}
A recent survey on AL can be found in the work of Tharwat \etal \cite{AL_survey} and a survey on AL applied to the medical domain \cite{medical_AL_survey}.
AL methods can be categorized into three groups: uncertainty-based \cite{entropy,MC_dropout,BALD,BALD_non_img,feature_mixing}, diversity-based \cite{core_set, vae_AL}, and methods that combine both uncertainty and diversity \cite{suggestive_ann,BADGE,contrastive_AL,MEAL,batchBALD}. Our approach falls into the last category and proposes to select samples based on their diversity and the uncertainty of the model.

In the same category as our method, Ash \etal \cite{BADGE} compute the gradient embeddings of the model classification predictions and use the magnitude of these embeddings to assess the uncertainty of the corresponding predictions. They proceed to sample inputs whose gradient embedding is far from each other using the $k$-means++ initialization method. The gradient embedding is obtained by multiplying the predicted probability of the image with a hidden layer embedding, a vector. In the case of segmentation, we have a 2D map of probabilities instead of a single scalar. Computing the gradient embedding by multiplying the probability of each pixel with the embedding and then stacking them would yield gigantic vectors, hence it is not straightforward to adapt their method for segmentation tasks.

Margatina \etal \cite{contrastive_AL} get the embeddings for input text data and sample inputs whose classification predictions differ despite having similar embeddings. Their method assumes the prediction is a probability vector and can not be applied directly to segmentation tasks. 

Other methods, such as \cite{feature_mixing,batchBALD} also rely on the classification nature of the task, and it is not straightforward to adapt them for segmentation. We thus compare our approach to the following uncertainty-based methods \cite{entropy,MC_dropout,BALD}, diversity-based methods \cite{core_set,vae_AL} and uncertainty-diversity methods \cite{suggestive_ann}.
\section{Method}

% \begin{figure}[t]
%     \includegraphics[width=1\linewidth]{figures/auris_dataset.pdf}
%     \medskip
%     \includegraphics[width=1\linewidth]{figures/auris_seq_frames.pdf}
%     \medskip
%     \includegraphics[width=1\linewidth]{figures/intuitive_dataset.pdf}
%     \medskip
%     \includegraphics[width=1\linewidth]{figures/intuitive_seq_frames.pdf}
%    \caption{First 4 rows show examples of images obtained from the MONARCH\texttrademark~ platform. Rows 1 and 2 show different instruments (REBUS sheath, needle clear sheath, forceps sheath, needle tip, needle blue sheath, needle brush, forceps clamp, and REBUS probe) in the dataset and their corresponding segmentation masks. Row 2 and 4 show partial image sequences at 2~fps. The last 4 rows show examples of images from the EndoVis \cite{endovis} dataset. Rows 5 and 6 show different tools (instrument shaft, instrument clasper, instrument wrist, thread, clamps, needle, suction instrument, and ultra-sound probe) in the dataset and their corresponding segmentation masks. Row 7 and 8 show partial image sequence at 1~fps.}
% \label{fig:auris}
% \end{figure}

\begin{figure}[t]
    \includegraphics[width=1\linewidth]{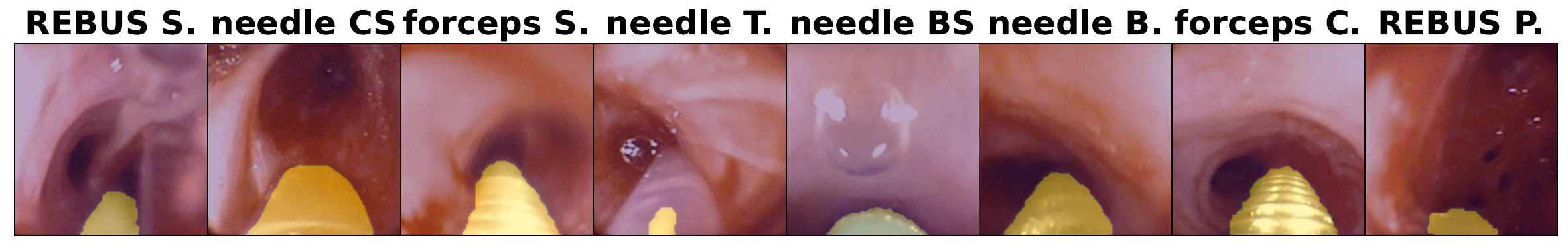}
    \medskip
    \includegraphics[width=1\linewidth]{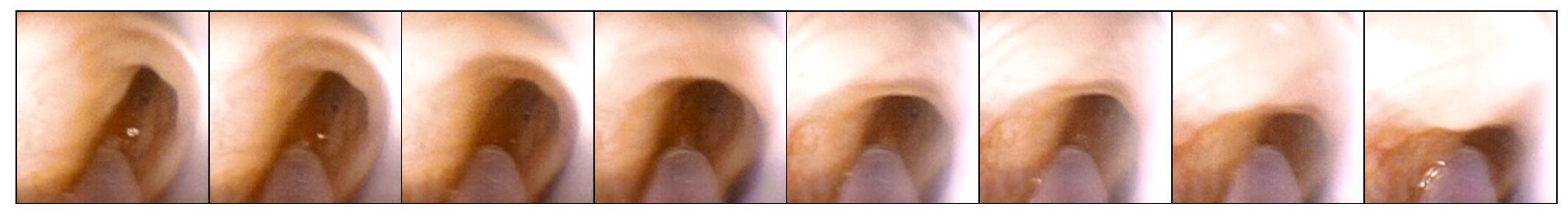}
    \medskip
    \includegraphics[width=1\linewidth]{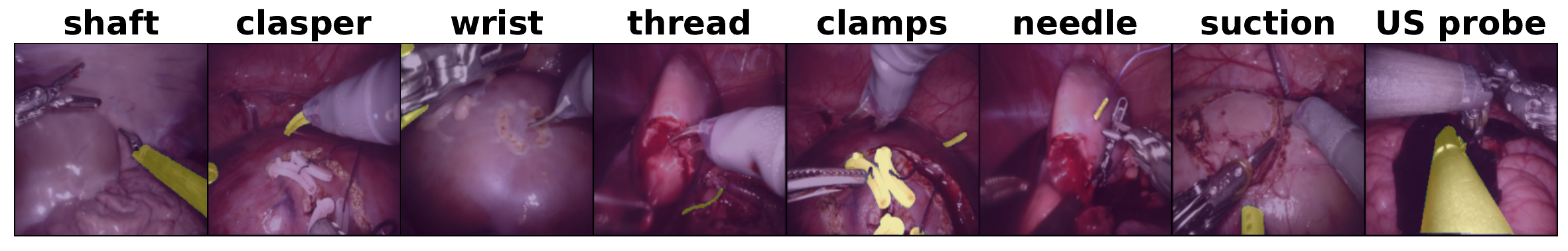}
    \medskip
    \includegraphics[width=1\linewidth]{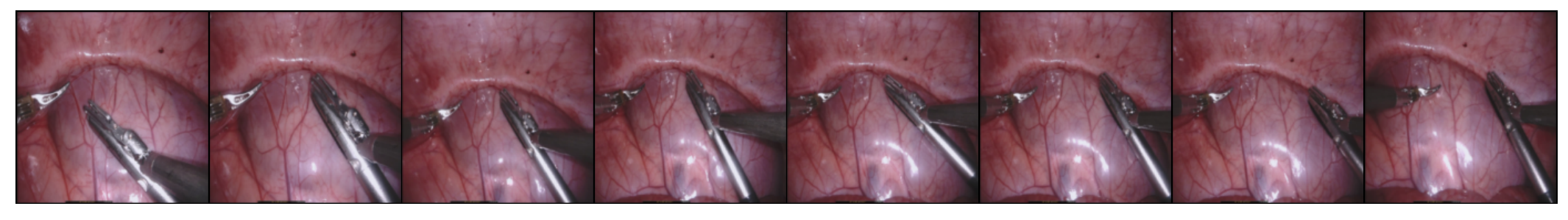}
    \medskip
    \includegraphics[width=1\linewidth]{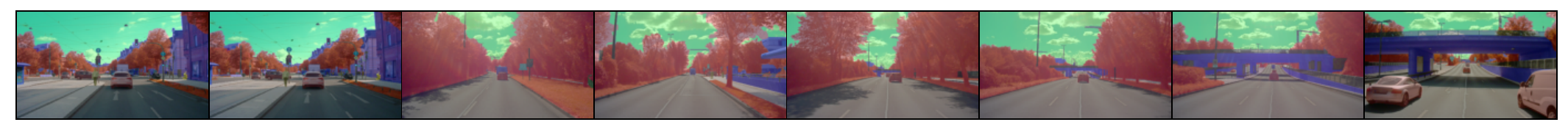}
    \medskip
    \includegraphics[width=1\linewidth]{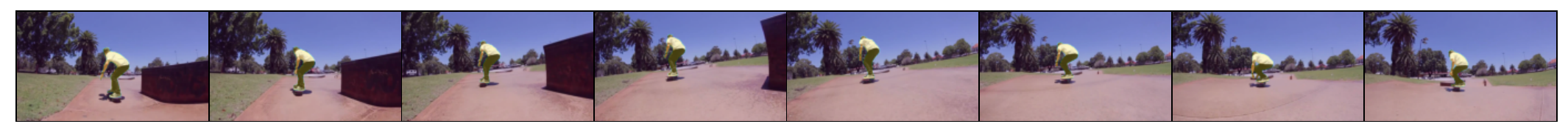}
    \medskip
    \includegraphics[width=1\linewidth]{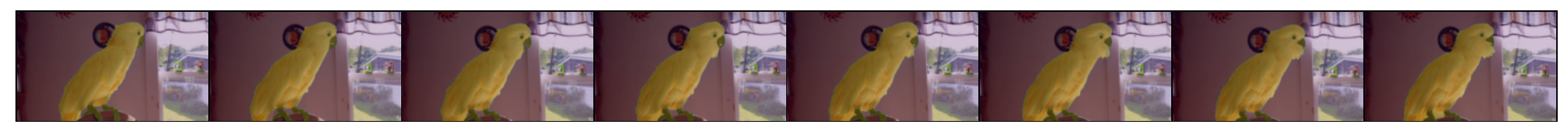}
   \caption{First 2 rows show examples of images obtained from the MONARCH\texttrademark~ Platform. Row 1 shows different instruments (various sheaths, needle, brush, forceps, REBUS probe) in the dataset and their corresponding segmentation masks. Row 2 shows partial image sequences at 2~fps. Row 3 and 4 show examples of images from the EndoVis \cite{endovis} dataset. Row 3 shows different tools (instrument shaft, instrument clasper, instrument wrist, thread, clamps, needle, suction instrument, and ultra-sound probe) in the dataset and their corresponding segmentation masks. Row 4 partial image sequence at 1~fps. Row 5 shows partial image sequences of the A2D2 \cite{A2D2} dataset. Row 6 and 7 show image sequences of the Skateboard and Parrot datasets \cite{YouTube-VOS} at 6~fps.}
\label{fig:auris}
\end{figure}

A video segmentation model is a function~$f:\cX\to\cY$ that receives an image, or video frame, $\x\in\cX$ and produces a label map $\y=f(\x)\in\cY$. Training such a model requires pairs $(\x, \y)\in\cX\times\cY$ of video frames and their corresponding annotated label maps. While obtaining video frames is usually inexpensive, producing manual annotations is time-consuming and effort-intensive. Active learning is a technique that reduces the number of required annotations by iteratively selecting the most informative frames for annotation. At each step~$t$, a subset of the unlabeled frames is selected according to an \emph{annotation strategy function}~$\pi:2^\cX\times 2^{\cX\times\cY}\to 2^\cX$ that receives the set of unlabeled frames~$\U_t$ and the set of labeled frames~$\A_t$ and suggests a subset of frames~$\pi(\U_t, \A_t)\subset\U_t$ that should be annotated. After annotation, these frames are moved to the set of labeled frames for the next step~$\A_{t+1} \subset \cX\times\cY$. The model is then trained with~$\A_{t+1}$, and the whole process is repeated in the next step.

In this work, we focus on the problem of active learning with video sequences. Given a collection of unlabelled videos~$\{\X^{(v)}\}_{v=1}^V$, where each video is a sequence of frames~$\X^{(v)} = (\x^{(v)}_1, \ldots, \x^{(v)}_{F_v})$, our goal is to find a frame annotation strategy~$\pi$ that maximizes the performance of the segmentation model~$f$ for a fixed annotation budget. This problem is challenging since video frames are not independent and identically distributed (\emph{i.i.d.}),~ a default assumption in most active learning methods. While it is possible to treat the set of frames as \emph{i.i.d.}~samples, in practice, video frames exhibit strong temporal dependencies that we can exploit in the design of~$\pi$.

\subsection{Annotation strategy}
Our strategy is to select unlabeled frames representing highly correlated video segments while avoiding selecting redundant frames similar to those already in the labeled dataset. At each step~$t$, the policy first projects all frames in $\A_t$ and~$\U_t$ to a representation space~$\cZ$ with an embedding function~$\phi: \cX\to\cZ$. K-means are applied in this space with $K=|\A_t| + Q$, where $|\cdot|$ is the cardinal of a set, and $Q$ is a hyperparameter indicating the desired number of selected frames at each step. This process finds a set~$\{\k^{(i)}\}_{i=1}^K$ of representative centroids in the embedding space. We then look for centroids similar to the frames in~$\A_t$ in terms of distance in the embedding space.

% Specifically, we use the Hungarian algorithm 
We design a matching algorithm (see Appendix)
to find the matching~$\{(\a^{(i)}, \k^{(m_i)})\}_{i=1}^{|\A_t|}$ between the centroids~$\mathbf{k}$ and the embeddings~$\mathbf{a}$ of the elements of~$\A_t$. The distance between a centroid~$\mathbf{k}$ and~$\A_t$ is defined as $\min_{\mathbf{a} \in \mathbf{A}_t} \|\mathbf{a} - \mathbf{k}\|$.
We first assign the centroids~$\mathbf{k}$ with the smallest distances to~$\A_t$ with the corresponding embedding~$\mathbf{a}$ used in the distance calculation.
Since we have more centroids than frames in $\A_t$, this matching ensures that $Q$~centroids with the highest distances to $\A_t$ are left out. Each matched centroid~$\k^{(m_i)}$ is substituted by its corresponding vector~$\a^{(i)}$. 

A second round of k-means is applied to update the remaining $Q$~unmatched centroids while keeping the matched centroids fixed to their new values~$\a^{(i)}$. This process ensures that the new $Q$~centroids will represent video segments not already covered by the labeled samples since labeled samples belong to the clusters of the fixed centroids. Finally, the selected samples by our strategy are those with the highest entropy \cite{endovis} from each cluster of the $Q$~unmatched centroids. The entropy of a frame is computed as the sum of pixel entropies, which are computed in a standard way by applying the entropy formula to the output probabilities of the segmentation model. The probabilities are given after the softmax or sigmoid operation.

Note that just running k-means once and fixing $|\A_t|$~centroids to the corresponding embeddings of labeled frames is highly dependent on the initialization of the remaining~$Q$ centroids, and we found experimentally to lead to suboptimal solutions, as the fixed centroids prevent the others from moving and adequately covering the embedded vectors.

Our strategy~$\pi$ assumes that the distance between embedding vectors represents the level of correlation between frames. In particular, the distance between highly correlated frames should be smaller than between independent frames. We design our embedding function accordingly.

\subsection{Representation learning}
\label{sect:simclr}
 The embedding function~$\phi:\cX \to \cZ$ is modeled as a ResNet-34~\cite{resnet} trained in an unsupervised way with SimCLR~\cite{simCLR} once before AL starts. In particular, SimCLR minimizes the contrastive loss
 \begin{equation}
 \label{eq:contrastive_loss}
 \ell_{i, j}=-\log\frac{\exp\left(\simil(\z_i,\z_j)/\tau\right)}{\sum_{k=1}^{2N}\mathbbm{1}_{[k\neq i]}\exp\left(\simil(\z_i,\z_k)/\tau\right)}
 \end{equation}
 where~$\simil$ is the cosine similarity $\simil(\mathbf{u}, \mathbf{v}) = \frac{\mathbf{u}^{T}\cdot \mathbf{v}}{\| \mathbf{u} \|\cdot \| \mathbf{v} \|}$, $\z_i=\phi(\x_i)$. The loss is minimized for all pairs of frames augmented from the same image.

\section{Experiments}

% \begin{figure*}[t]
%     \includegraphics[width=0.5\linewidth]{figures/auris_results_resnet50.pdf}
%     \includegraphics[width=0.5\linewidth]{figures/intuitive_results_resnet50.pdf}
%     \medskip
%     \includegraphics[width=0.5\linewidth]{figures/auris_results_resnet101.pdf}
%     \includegraphics[width=0.5\linewidth]{figures/intuitive_results_resnet101.pdf}
%    \caption{Comparison of different sampling strategies in terms of test set DICE coefficient. Experiments are conducted on the MONARCH and EndoVis datasets with the ResNet-50 or ResNet101 backbones. Evaluations are repeated for 10 different training-validation splits. Error bars indicate one standard error.}
% \label{AL_results}
% \end{figure*}

\begin{figure*}[t]
    \includegraphics[width=0.5\linewidth]{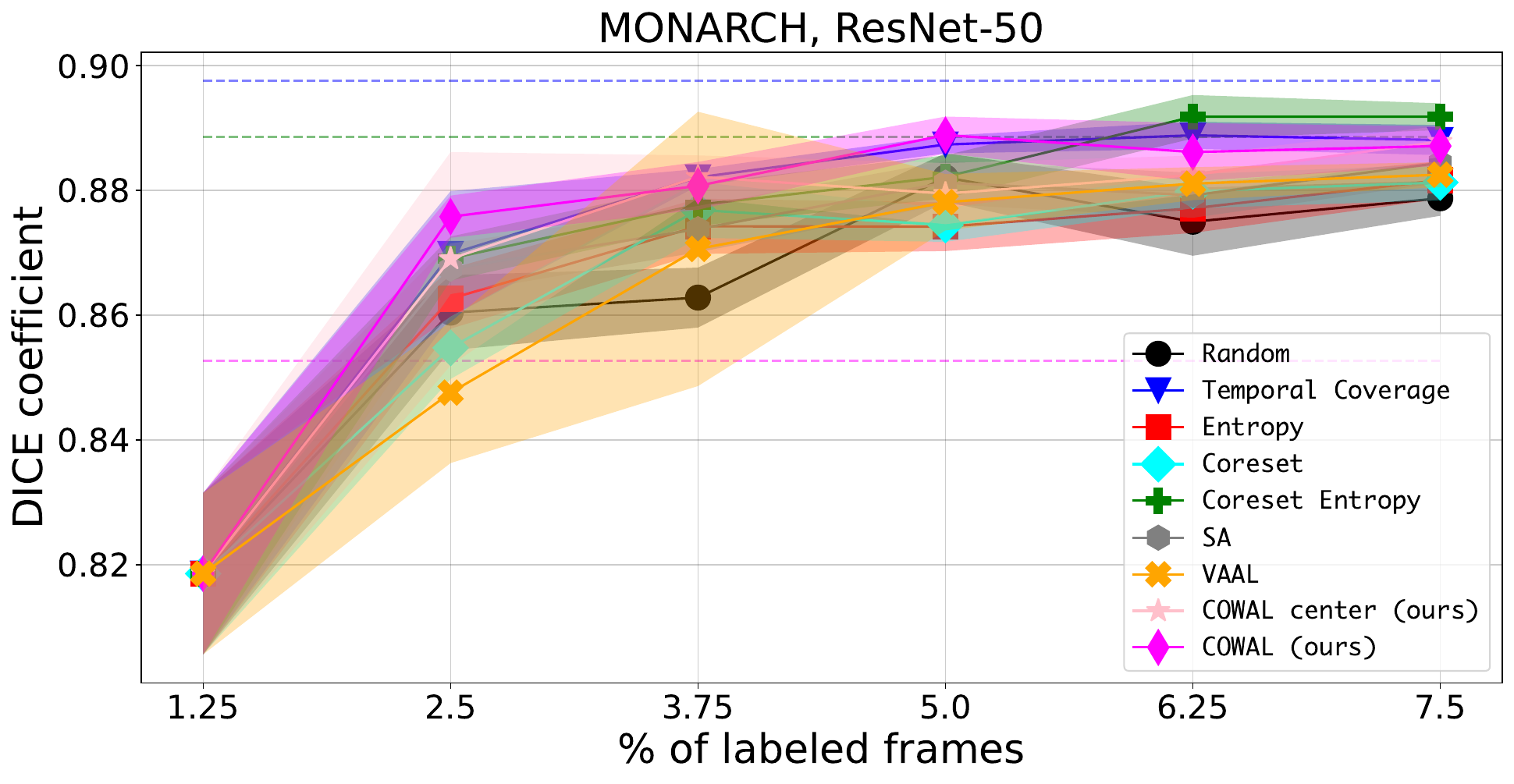}
    \includegraphics[width=0.5\linewidth]{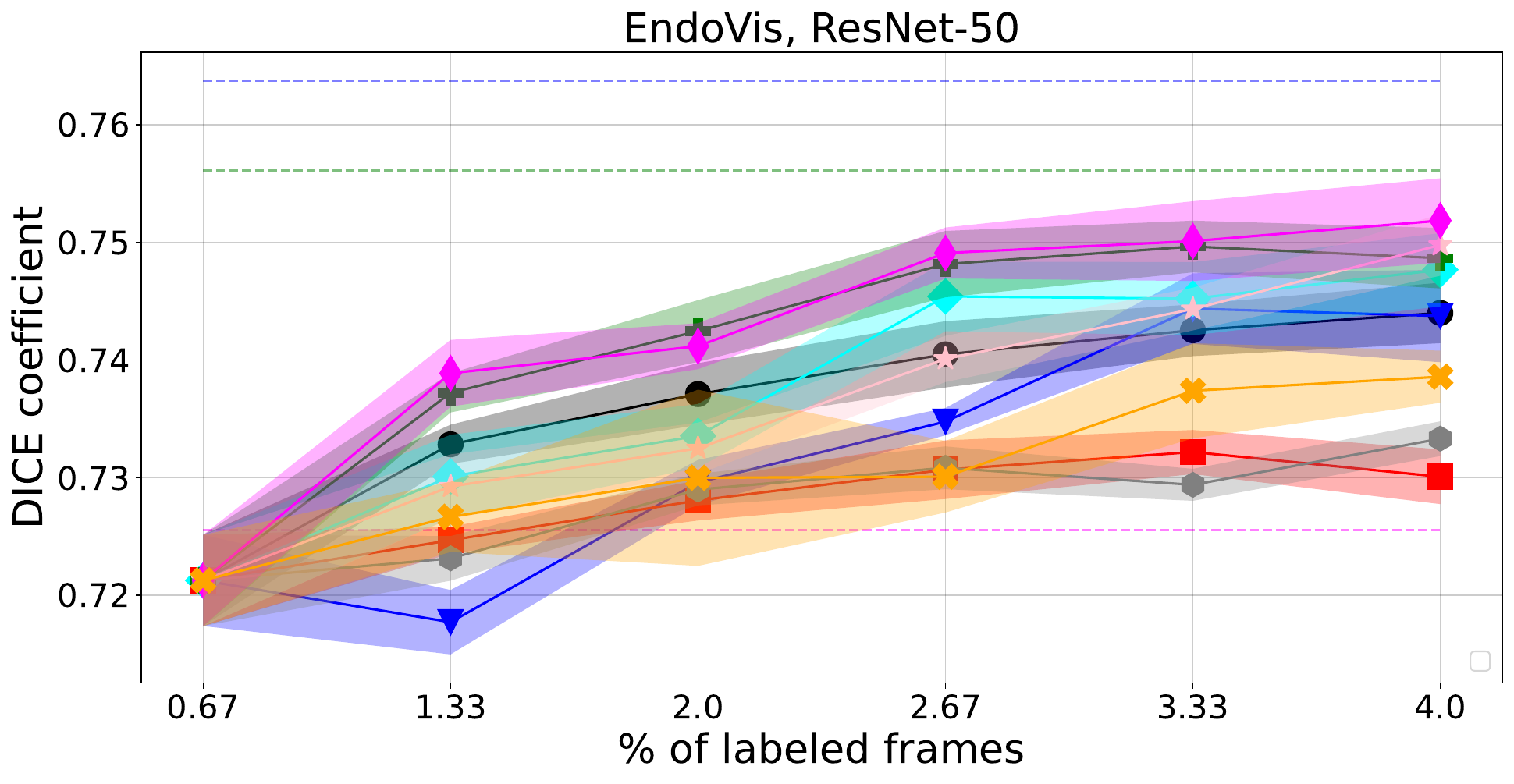}
    \medskip
    \includegraphics[width=0.5\linewidth]{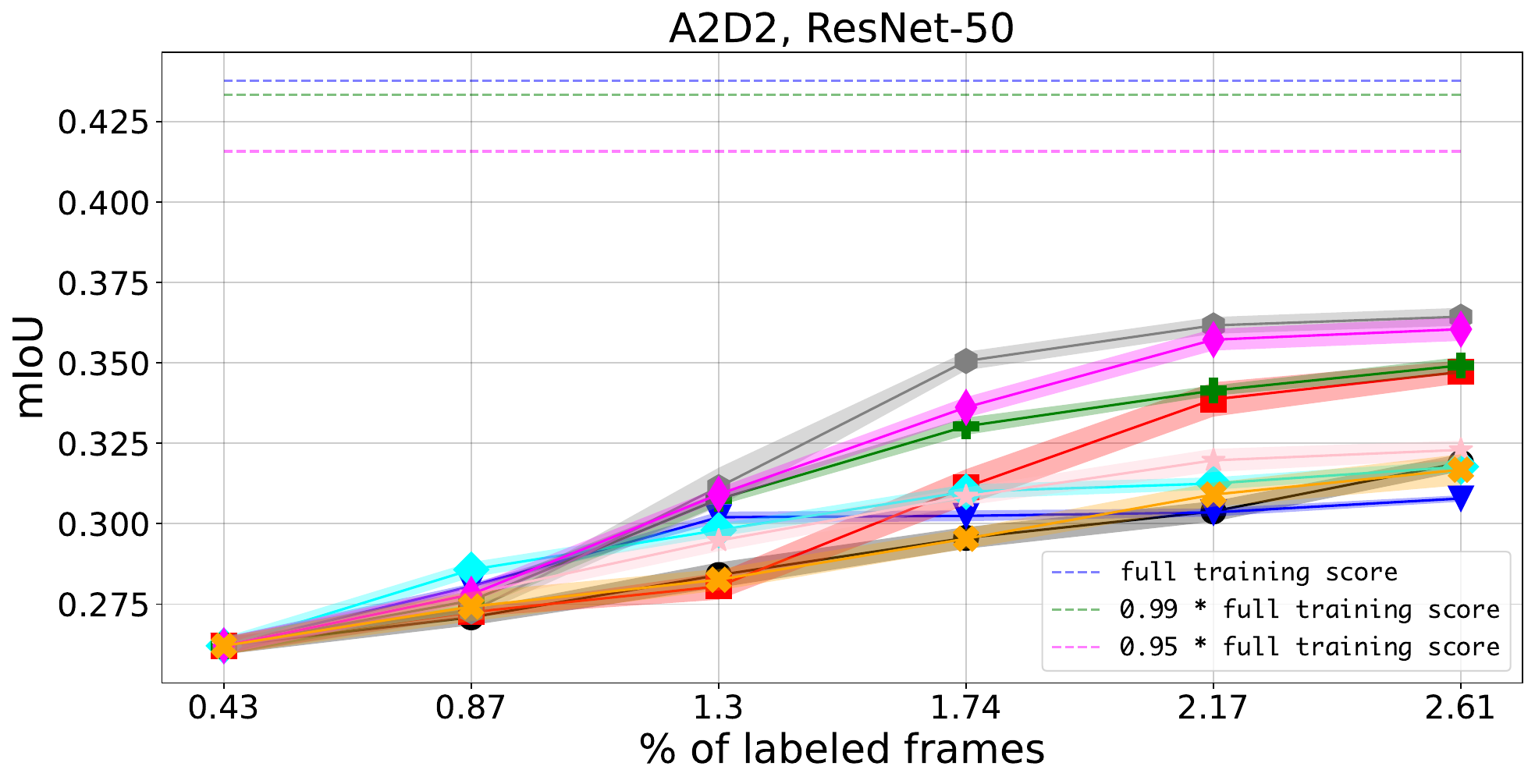}
    \includegraphics[width=0.5\linewidth]{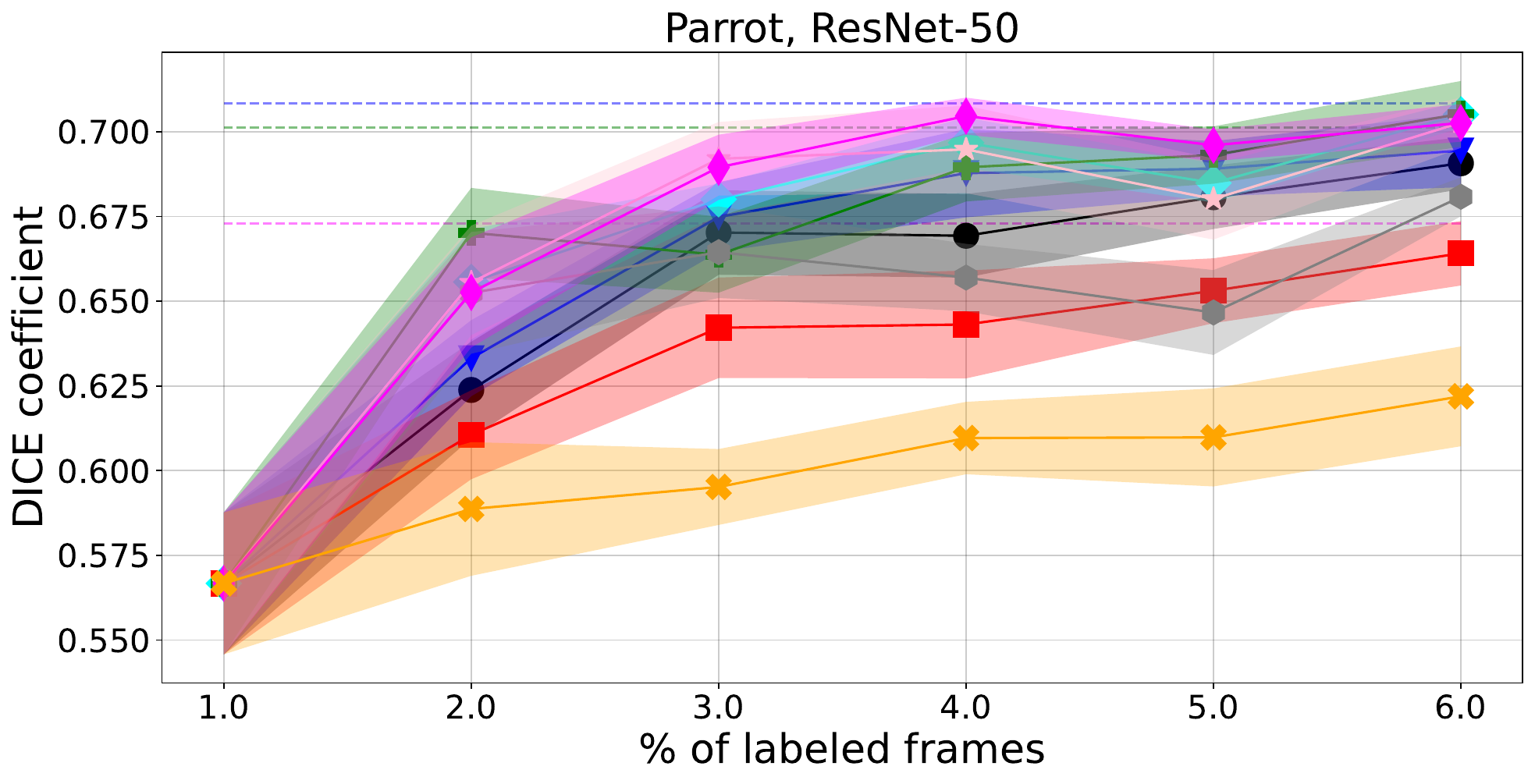}
   \caption{Comparison of different sampling strategies.
   % Experiments were conducted on the MONARCH, EndoVis \cite{endovis}, A2D2 \cite{A2D2}, Skateboard \cite{YouTube-VOS}, and Parrot \cite{YouTube-VOS} datasets with the ResNet-50 or ResNet101 backbones \PMN{Only ResNet50 results are shown?}.
   Values are averaged over 10 different training-validation splits, with error bars indicating one standard deviation. Plots for Skateboard and ResNet101 backbones are available in Appendix~A.}
\label{AL_results}
\end{figure*}

\subsection{Dataset} 
\textbf{MONARCH}. We collected and created a video dataset using the MONARCH{\texttrademark} Platform, a robotic-assisted bronchoscopy system indicated for diagnostic and therapeutic procedures within the lung. The dataset centers on the biopsy phase, encompassing the insertion of diverse instruments such as REBUS, needles, forceps, and brushes. It contains 11~videos of bronchoscopy procedures with segmentation masks on the biopsy tools acquired following HIPAA requirements. These videos were subdivided into 56 segments, collectively amounting to 2883 frames. Segments devoid of visible biopsy tools were excluded from the experiments.

The videos were recorded at 25~fps with a resolution of~$220\times{}220$, and the segmentation annotation were done at 2~fps with a labeled frame every 12~frames. Examples of the different tools and their segmentation mask are displayed in Fig.~\ref{fig:auris}. In this dataset, only one tool is visible per frame. We separate the 56 videos into 30~videos for training ($\approx1'600$~frames) and 26~videos for testing ($\approx 1'300$~frames). For our experiments, the dataset is simplified from a multi-tool segmentation task to a binary segmentation task with biopsy tools versus background annotations.

% \begin{table}
% \centering
% \scalebox{0.9}{
% \begin{tabular}{c | c | @{\hspace{0cm}} c | c}
% Sampling Method & Backbone & \quad MONARCH & EndoVis \\
% \hline
% Random & ResNet50 & \quad 0.804 & 0.804 \\
% Temporal Coverage & ResNet50 & \quad 0.813 & 0.799 \\
% Entropy \cite{entropy} & ResNet50 & \quad 0.805 & 0.795 \\
% CoreSet \cite{core_set} & ResNet50 & \quad 0.805 & 0.805 \\
% SA \cite{suggestive_ann} & ResNet50 & \quad 0.808 & 0.794 \\
% VAAL \cite{vae_AL} & ResNet50 & \quad 0.804 & 0.797 \\
% COWAL (ours) & ResNet50 & \quad \textbf{0.814} & \textbf{0.811} \\
% \hline
% Random & $\text{ResNet101}^*$ & \quad 0.815 & 0.809 \\
% Temporal Coverage & $\text{ResNet101}^*$ & \quad 0.817 & 0.803 \\
% Entropy \cite{entropy} & $\text{ResNet101}^*$ & \quad 0.810 & 0.799 \\
% CoreSet \cite{core_set} & $\text{ResNet101}^*$ & \quad 0.815 & 0.811 \\
% SA \cite{suggestive_ann} & $\text{ResNet101}^*$ & \quad 0.814 & 0.801 \\
% VAAL \cite{vae_AL} & $\text{ResNet101}^*$ & \quad 0.811 & 0.802 \\
% MC Dropout \cite{MC_dropout} & $\text{ResNet101}^*$ & \quad 0.810 & 0.800 \\
% BALD \cite{BALD} & $\text{ResNet101}^*$ & \quad 0.805 & 0.806 \\
% COWAL (ours) & $\text{ResNet101}^*$ & \quad \textbf{0.819} & \textbf{0.816} \\
% \hline
% \end{tabular}}
% \caption{AuALC of all sampling methods across all datasets. Metrics are computed at the end of the active learning steps. The $*$ in $\text{ResNet101}^*$ indicates that it is a Bayesian model which allows the use of sampling methods such as MC Dropout \cite{MC_dropout} or BALD \cite{BALD}.}
% \label{table_result}
% \end{table}

\begin{table*}
\centering
\begin{tabular}{c | c | @{\hspace{0cm}} c | c | c | c | c | c}
Sampling Method & Backbone & \quad MONARCH & EndoVis & A2D2 & Skateboard & Parrot & Avg across datasets\\
\hline
Random & ResNet50 & \quad 0.804 & 0.804 & 0.550 & 0.790 & 0.770 & 0.744\\
Temporal Coverage & ResNet50 & \quad \underline{0.813} & 0.799 & 0.561 & 0.746 & 0.780 & 0.740\\
Entropy \cite{entropy} & ResNet50 & \quad 0.805 & 0.795 & 0.574 & 0.768 & 0.745 & 0.737\\
CoreSet \cite{core_set} & ResNet50 & \quad 0.805 & 0.805 & 0.570 & 0.789 & 0.789 & 0.752\\
CoreSet x Entropy & ResNet50 & \quad \underline{0.813} & \underline{0.810} & 0.594 & 0.781 & 0.789 & \underline{0.757}\\
SA \cite{suggestive_ann} & ResNet50 & \quad 0.808 & 0.794 & \textbf{0.613} & 0.769 & 0.763 & 0.749\\
VAAL \cite{vae_AL} & ResNet50 & \quad 0.804 & 0.797 & 0.552 & 0.740 & 0.705 & 0.720\\
COWAL center (ours) & ResNet50 & \quad 0.811 & 0.803 & 0.569 & \textbf{0.799} & \underline{0.790} & 0.754\\
COWAL (ours) & ResNet50 & \quad \textbf{0.814} & \textbf{0.811} & \underline{0.606} & \underline{0.793} & \textbf{0.795} & \textbf{0.764}\\
\hline
Random & $\text{ResNet101}^*$ & \quad 0.815 & 0.809 & 0.584 & \underline{0.804} & 0.818 & 0.766\\
Temporal Coverage & $\text{ResNet101}^*$ & \quad \underline{0.817} & 0.803 & 0.579 & 0.776 & \underline{0.846} & 0.764\\
Entropy \cite{entropy} & $\text{ResNet101}^*$ & \quad 0.810 & 0.799 & 0.589 & 0.785 & 0.781 & 0.753\\
CoreSet \cite{core_set} & $\text{ResNet101}^*$ & \quad 0.815 & 0.811 & 0.590 & 0.802 & 0.831 & 0.770\\
CoreSet x Entropy & $\text{ResNet101}^*$ & \quad \underline{0.817} & \underline{0.812} & 0.616 & 0.787 & 0.835 & \underline{0.773}\\
SA \cite{suggestive_ann} & $\text{ResNet101}^*$ & \quad 0.814 & 0.801 & \textbf{0.632} & 0.790 & 0.797 & 0.767\\
VAAL \cite{vae_AL} & $\text{ResNet101}^*$ & \quad 0.811 & 0.802 & 0.586 & 0.763 & 0.759 & 0.744\\
MC Dropout \cite{MC_dropout} & $\text{ResNet101}^*$ & \quad 0.810 & 0.800 & . & . & . & .\\
BALD \cite{BALD} & $\text{ResNet101}^*$ & \quad 0.805 & 0.806 & . & . & . & .\\
COWAL center (ours) & $\text{ResNet101}^*$ & \quad 0.812 & 0.810 & 0.581 & \textbf{0.806} & 0.835 & 0.769\\
COWAL (ours) & $\text{ResNet101}^*$ & \quad \textbf{0.819} & \textbf{0.816} & \underline{0.621} & \underline{0.804} & \textbf{0.848} & \textbf{0.782}\\
\hline
\end{tabular}
\caption{AuALC of all sampling methods across all datasets. Metrics are computed at the end of the active learning steps. The $*$ in $\text{ResNet101}^*$ indicates that it is a Bayesian model which allows the use of sampling methods such as MC Dropout \cite{MC_dropout} or BALD \cite{BALD}. Best scores are in bold and second best are underlined.}
\label{table_result}
\end{table*}

\textbf{EndoVis} \cite{endovis}. We also evaluate our method on the EndoVis 2018 Robotic Scene Segmentation dataset \cite{endovis}. It is a dataset of surgery videos with segmentation masks on surgery tools and human body organs. A total of 19 videos are divided into 15 training videos ($\approx$ 2250 frames) and 4 test videos ($\approx$ 1000 frames). Each video came from a single porcine training procedure. Images from the left eye in the stereo pair are used for training. The annotation procedure is done at 1 fps with non-anatomical classes such as instrument shaft, instrument wrist, instrument clasper, threads, clamps, suturing needle, suction instrument, and ultra-sound probe. An example of each tool is shown in Fig.~\ref{fig:auris} row~3. Like the MONARCH dataset, we simplified the annotations with an anatomical vs.~non-anatomical binary mask.

\textbf{A2D2} \cite{A2D2} is a large-scale driving dataset consisting of 41277 annotated images with a resolution of~$1208\times1920$ from 23 sequences. It covers an urban setting from highways, country roads, and three cities. It contains labels for 38 categories. For our experiments, we map them to the 19 classes of Cityscapes. Following the procedure of \cite{best_practices_AL}, we extract 60 video segments from the original A2D2 dataset. Each segment contains 44 consecutive frames, totaling 2640 frames. The video \textit{'20180925\_112730'} with 993 frames is used as the test set. The annotation rate is irregular, varying between 10 to 300~frames per annotation.

\textbf{YouTube-VOS} \cite{YouTube-VOS} is the most extensive video segmentation dataset with more than 5000 videos and 90 classes. We selected only videos showing skateboarders or parrots to build two binary segmentation datasets. The Skateboard dataset contains 24 videos ($\approx$ 700 frames) for training and 10 videos ($\approx$ 250 frames) for testing. The Parrot dataset contains 46 videos ($\approx$ 1500 frames) for training and 10 videos ($\approx$ 500 frames) for testing.

\subsection{Baselines}
We evaluate \COALSamp{} against the following baselines methods, \RSamp, \ESamp~\cite{entropy}, \MCSamp~\cite{MC_dropout}, \BALDSamp~\cite{BALD}, \SASamp~\cite{suggestive_ann}, \CoreSamp~\cite{core_set}, \VAALSamp~\cite{vae_AL} as well as some custom designed baselines:\\
- \textbf{\TempSamp} selects samples by prioritizing videos with fewer labeled frames, and within a video, select the temporally most distant frame to currently labeled frames. \\
- \textbf{\CoreEntSamp} computes the distance of an unlabeled sample to the labeled set as in \CoreSamp \cite{core_set} and scale these distances by the entropy values~\cite{entropy}, selecting the samples with the highest scaled distances. We use the same embedding~$\phi$ as \COALSamp{}.\\
- \textbf{\COALCenterSamp}, we also compare \COALSamp~with an ablated version of itself. Instead of selecting the frame with the highest entropy per cluster, we select the frame closest to the centroid to maximize sample diversity.

\subsection{Implementation details}
We follow the same experimental setup for all the baselines.
The MONARCH input images are~$220\times220$, the EndoVis images are ~$224\times224$, the A2D2 images are~$270\times480$, and the YouTube-VOS images are ~$256\times448$. For data augmentation, we take crops with random scale factors in the range~$(0.85, 1)$ of the original image and with random aspect ratios in the range~$(\frac{3}{4}, \frac{4}{3})$. All crops are re-scaled back to the size of the original image, followed by a random horizontal flipping with a probability of~$0.5$.

The set of labeled frames~$\A_1$ is initialized with the middle frame of 10~training videos and their corresponding annotations. The complementary set of unlabeled frames~$\U_1$~contains the remaining frames from the training data. At each step~$t$, the segmentation model~$f$ is trained with~$\A_t$ until convergence of the DICE~score on the validation set.
We run each baseline 10~times, randomly splitting the training videos into training and validation sets with a proportion of~$2:1$ for the EndoVis dataset,~$1:1$ for the MONARCH,~$9:1$ for A2D2, ~$2:1$ for Parrot, and~$3:1$ for Skateboard. The reason for a~$1:1$ split for MONARCH is that the videos from this dataset are usually shorter and contain many redundant frames. Hence, having many videos in the validation set ensures it contains diversified images.
The segmentation model~$f$ is a DeepLabV3~\cite{deeplabV3} for which we have two different versions. The first version is from the official PyTorch/vision GitHub repository~\cite{deeplab_pytorch} and uses the ResNet-50 backbone. The second version, from the implementation of~\cite{viewAL}, uses a ResNet-101 backbone and a deeper decoder with more dropout layers. % Aside from the backbone, the decoder between both versions is also different, as the second version has a deeper decoder with more dropout layers to approximate a Bayesian behavior. 

The training is done with a batch size of~4, a learning rate of~$10^{-4}$, and no weight decay using the Adam optimizer~\cite{adam}. The patience is set to~20, and at every AL step, evaluation on the validation set is done after the number of iterations required to go through the whole training set. We apply Polyak averaging with~$\alpha=0.99$ for stable training evolution. At the first AL step $t=1$, the model is initialized with ImageNet pre-trained weights and, for subsequent steps, using the weights obtained at the end of the previous step. The policy function chooses~$Q=10$~new samples for annotation at each AL step. After convergence, model performance is evaluated with the DICE~score on the test set. 
% \PMN{I do not think these time measurements are very useful. Remove?}. \fei{I read in the reproducibility section of the paper submission guide that providing run time is good, but we can remove it}. It takes around 20~hours on a single RTX~3090Ti to obtain the 10 runs results of one sampling method for the MONARCH dataset, 60 hours for EndoVis, and 60 hours for A2D2.On a single GTX~1080Ti, it takes around 48 hours for Skateboard, and 72 hours for Parrot.

We train the embedding function~$\phi$ with contrastive learning on the train and validation sets of the respective dataset as described in Sect.~\ref{sect:simclr}. We used a learning rate of~$3\cdot10^{-4}$, a batch size of~256, a weight decay of~$10^{-2}$, and trained for 2000~epochs using the Adam optimizer~\cite{adam}. 

\subsection{Evaluation protocol}

We run each baseline 10~times, and perform 6~AL~steps on each run. We compute the median of the DICE scores over the 10~runs for each AL step and plot the resulting AL curves of \emph{median DICE} vs.~\emph{AL~step} to compare the performances. In addition, we report the area under the AL curve~(AuALC) as our evaluation metric. The AuALC is expressed as a fraction of the maximum possible area that a hypothetically perfect AL method would achieve if it could reach the same performance as training the model with the complete training dataset at every AL step.

\begin{figure*}[t]
    \includegraphics[width=1\linewidth]{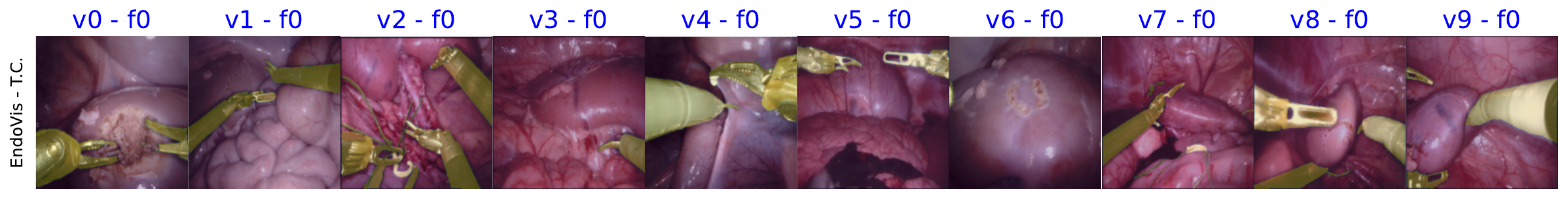}
    \includegraphics[width=1\linewidth]{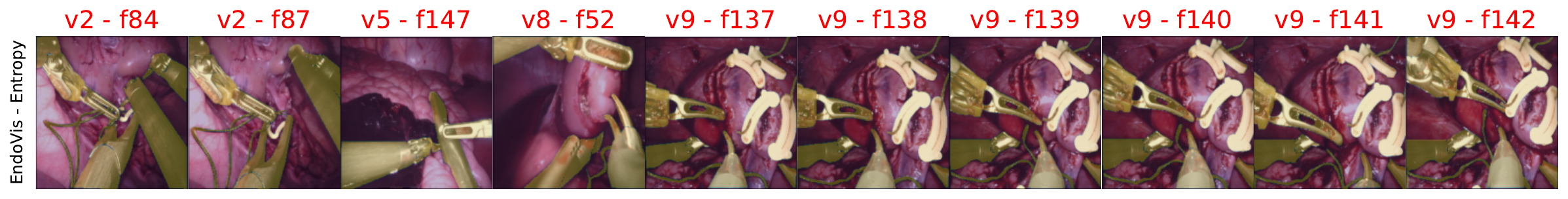}
    \includegraphics[width=1\linewidth]{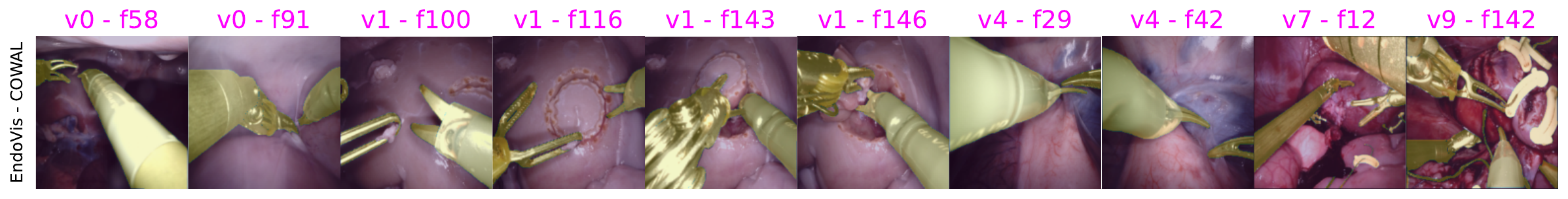}
    \includegraphics[width=1\linewidth]{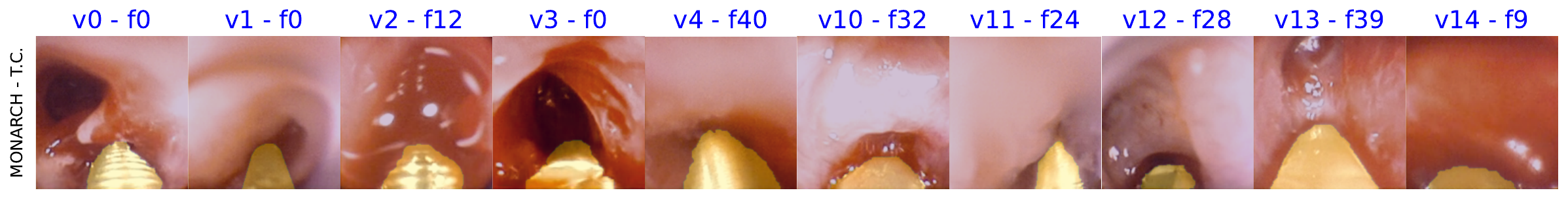}
    \medskip
    \includegraphics[width=1\linewidth]{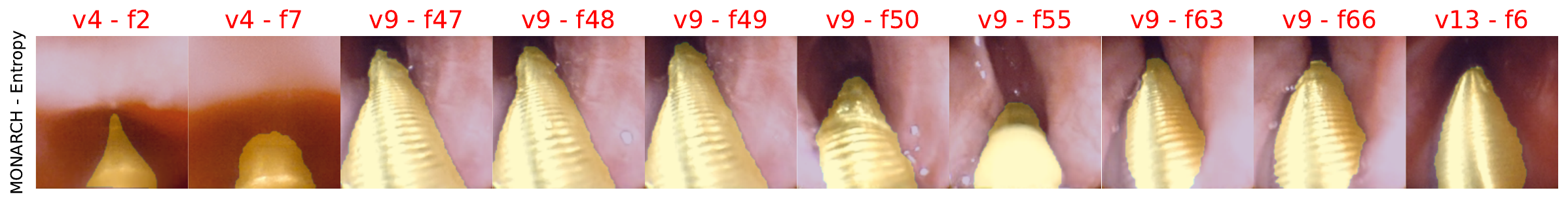}
    \includegraphics[width=1\linewidth]{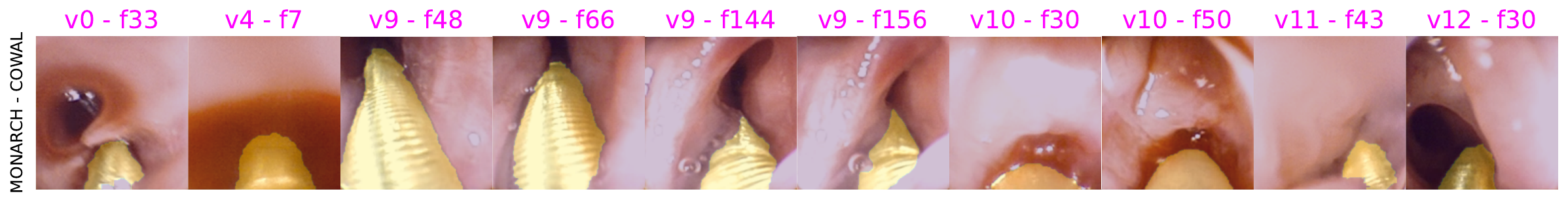}

   \caption{
   Selected frames at the first iteration. Video and frame numbers are indicated on top of each image. 
   }
\label{fig:sampling_examples}
\end{figure*}

\section{Results}

% \subsection{Behavior of Baseline Methods}

Fig.~\ref{AL_results} shows the experimental results. The differences in performance among baselines are significant, as indicated by the standard errors. 

\COALSamp{} outperformed all baselines for 3 datasets out of 5 and places second in the remaining 2 datasets (Table~\ref{table_result}). Strategies that rely only on uncertainty, such as~\ESamp{} \cite{entropy}, \MCSamp{} \cite{MC_dropout}, and \BALDSamp{} \cite{BALD}, performed poorly across all different settings in our experiments. These strategies have proven highly effective for image datasets \cite{MNIST, ISIC}, but they are more likely to select redundant frames when querying batches of frames from video sequences, as shown in Fig.~\ref{fig:sampling_examples}. For example, \ESamp{} selected multiple consecutive frames of video~9. This happens because frames that are temporally close often have similar uncertainty scores, causing a concentration of high-uncertainty frames within temporal neighborhoods.  In contrast,  \COALSamp{}, by ensuring a degree of visual diversity among labeled frames, selected only one frame from the same video.

% . We hypothesize that the model uncertainty about a particular frame will likely extend to similar frames in its temporal vicinity, causing the strategy also to select them. Fig.~\ref{fig:sampling_examples} shows the frames selected by \ESamp{} after training on~$\A_1$. This validates our assumption as we observe that \ESamp{} selected highly correlated frames: frames 47 to 49 of video 9 in row 4 and frames 137 to 142 of video 9 in row 2. In contrast, \COALSamp{} selected frames with high entropy while ensuring a degree of visual diversity among the set of labeled frames. Fig.~\ref{fig:sampling_examples} row 3 shows that \COALSamp{} only selected one frame from video~9, \textbf{v9-f142}, as the local representative of this segment. Whereas \ESamp{} could not assess the similarity of frames from video~9 and selected frames that contained redundant information.

While \textbf{\TempSamp{}} performed well on MONARCH, it struggled during the first AL iterations on the EndoVis dataset. \TempSamp{} starts by selecting the frames from the beginning and end of each video sequence, which often lack relevant information. For example, in Fig.~\ref{fig:sampling_examples} Row~1, selected frames \textbf{v3-f0} and \textbf{v6-f0} contain minimal presence of surgery tools, negatively impacting performance in the initial AL iterations.

% Since \TempSamp{} by design will first sample frames on the video extremities where surgeries begin or end, it can sometimes collect outliers. Row 1 in Fig.~\ref{fig:sampling_examples} shows frames selected by \TempSamp{} at 1.33\% labeling in the EndoVis - ResNet50 setting. Frames such as \textbf{v3-f0} and \textbf{v6-f0} contain little to no presence of surgery tool. Because at this stage of the AL method, there are approximately 20~labeled frames, having two uninformative frames can have a significant impact on the model performance.

\textbf{\SASamp}~\cite{suggestive_ann} outperforms \COALSamp{} on the A2D2 dataset due to lower frame redundancy. \SASamp{} selects high-entropy samples and diversifies within them, while \COALSamp{} diversifies first and then identifies high-entropy frames. \SASamp{} generally prioritizes higher-entropy samples over \COALSamp{}, yet frame redundancy can limit its effectiveness. In A2D2, with fewer redundant high-entropy samples, \SASamp{} performs better. For similar reasons, \ESamp~\cite{entropy} surpasses \RSamp{} and other diversity-based methods exclusively in the A2D2 dataset. The selected \SASamp{} frames are listed in Appendix~D.

\subsection{Embedding Model Ablation}
We evaluate the impact of removing our contrastive embedding function~$\phi$ and using the embedding defined by the downstream segmentation model~$f$ instead, as it is done in~\cite{core_set, suggestive_ann}. 
% AL methods \cite{core_set, suggestive_ann} that rely on image embedding to perform diversity sampling often use the embedding of the downstream task model (\eg,~the segmentation model~$f$) instead of training an contrastive learning model. We evaluate the impact of removing our embedding function and use the embedding of the task model instead, for \COALSamp{}, \CoreSamp{}, and \SASamp{} instead of the embedding $\phi$ from Sect.~\ref{sect:simclr}.
As shown in Table~\ref{table_ablation}, the use of the embedding~$\phi$ yields better results for \COALSamp{} and \CoreSamp{}, while the task model embedding is helpful for \SASamp{}. AL curves for this ablation are in Appendix~B.
% Table~\ref{table_ablation} summarizes the difference between using both embeddings and see Appendix B for the AL curves. 

\SASamp{} performs diversity sampling on a subset $S$ of the unlabeled samples, unlike the other two methods, which use the whole unlabeled set. The subset $S$ has high entropy frames, which for video datasets means similar frames,  and we argue that the quality of the embedding method is less relevant when applied to a pool of similar images hence explaining the different behavior of \SASamp{} compared to the other two methods. 

% \PMN{Not sure if we need to show results of TME for all baselines or only for ours. We are not proposing the contrastive embedding as a general, standalone method, but as part of ours. I guess it is enough to show that, for our method, the contrastive embedding leads to better performance.}

\begin{table}
\centering
\scalebox{0.85}{
\begin{tabular}{c | c | @{\hspace{0cm}} c | c}
Sampling Method & Backbone & \quad MONARCH & EndoVis \\
\hline
CoreSet with TME \cite{core_set} & ResNet50 & \quad 0.718 & 0.795 \\
CoreSet \cite{core_set} & ResNet50 & \quad 0.718 & \textbf{0.805} \\
\hline
SA with TME \cite{suggestive_ann} & ResNet50 & \quad \textbf{0.723} & \textbf{0.796} \\
SA \cite{suggestive_ann} & ResNet50 & \quad 0.722 & 0.794 \\
\hline
COWAL with TME (ours) & ResNet50 & \quad 0.721 & 0.803 \\
COWAL (ours) & ResNet50 & \quad \textbf{0.727} & \textbf{0.811} \\
\hline
\hline
CoreSet with TME \cite{core_set} & $\text{ResNet101}^*$ & \quad 0.814 & 0.805 \\
CoreSet \cite{core_set} & $\text{ResNet101}^*$ & \quad \textbf{0.815} & \textbf{0.811} \\
\hline
SA with TME \cite{suggestive_ann} & $\text{ResNet101}^*$ & \quad 0.813 & \textbf{0.803} \\
SA \cite{suggestive_ann} & $\text{ResNet101}^*$ & \quad \textbf{0.814} & 0.801 \\
\hline
COWAL with TME (ours) & $\text{ResNet101}^*$ & \quad 0.817 & 0.812 \\
COWAL (ours) & $\text{ResNet101}^*$ & \quad \textbf{0.819} & \textbf{0.816} \\
\hline
\end{tabular}}
\caption{AuALC of embedding ablation across \CoreSamp{}, \SASamp{}, and \COALSamp{} sampling strategies. The embedding method of each approach uses either the Task Model Embedding (TME) or the embedding $\phi$.}
\label{table_ablation}
\end{table}
 
\subsection{Budget Size}
\begin{figure}[t]
    \includegraphics[width=1\linewidth]{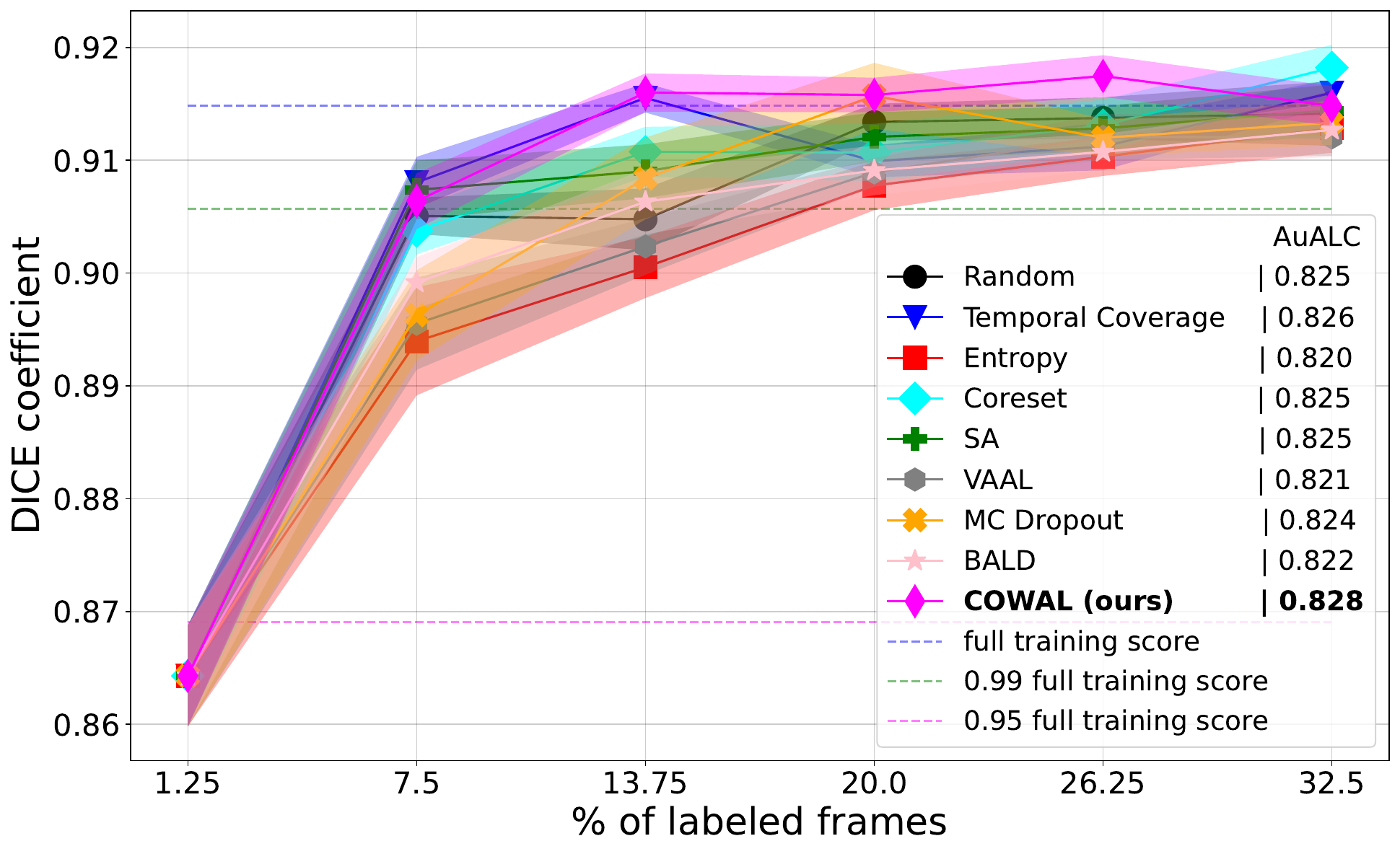}
   \caption{Comparison of different sampling strategies with a budget of $Q=50$ instead of 10. AuALC is displayed on the side.}
\label{budget_size_res}
\end{figure}

We compare sampling strategies with an increase selection size~($Q=50$) for the MONARCH-ResNet101 setting in Fig.~\ref{budget_size_res}. \COALSamp{} continues to outperform the other baselines. Among them, uncertainty-driven approaches, such as \ESamp{}~\cite{entropy} and \BALDSamp{}~\cite{BALD}, exhibit noticeably lower performance.

%-------------------------------------------------------------------------

\section{Conclusion}
We present a novel AL approach for video segmentation that aims to sample images that are diverse from previously selected frames, and that are uncertain according to the model. We select maximal entropy frames for each cluster, yielded by modified iterated k-means that enforce previously selected frames to differ from new ones. We experimentally show the effectiveness of our approach against different AL selection schemes, whereby our approach does consistently better. We visually show that our approach indeed provides a good strategy to diversify labeled images, by still selecting informative samples.
% which images should be sampled, by still selecting images that can improve the model effectively.

\section*{Acknowledgement}
This work is partially sponsored by Auris Health, Inc. (part of Johnson \& Johnson MedTech).
\begin{appendices}

\section{Additional plots}

\begin{figure*}
    \includegraphics[width=0.49\linewidth]{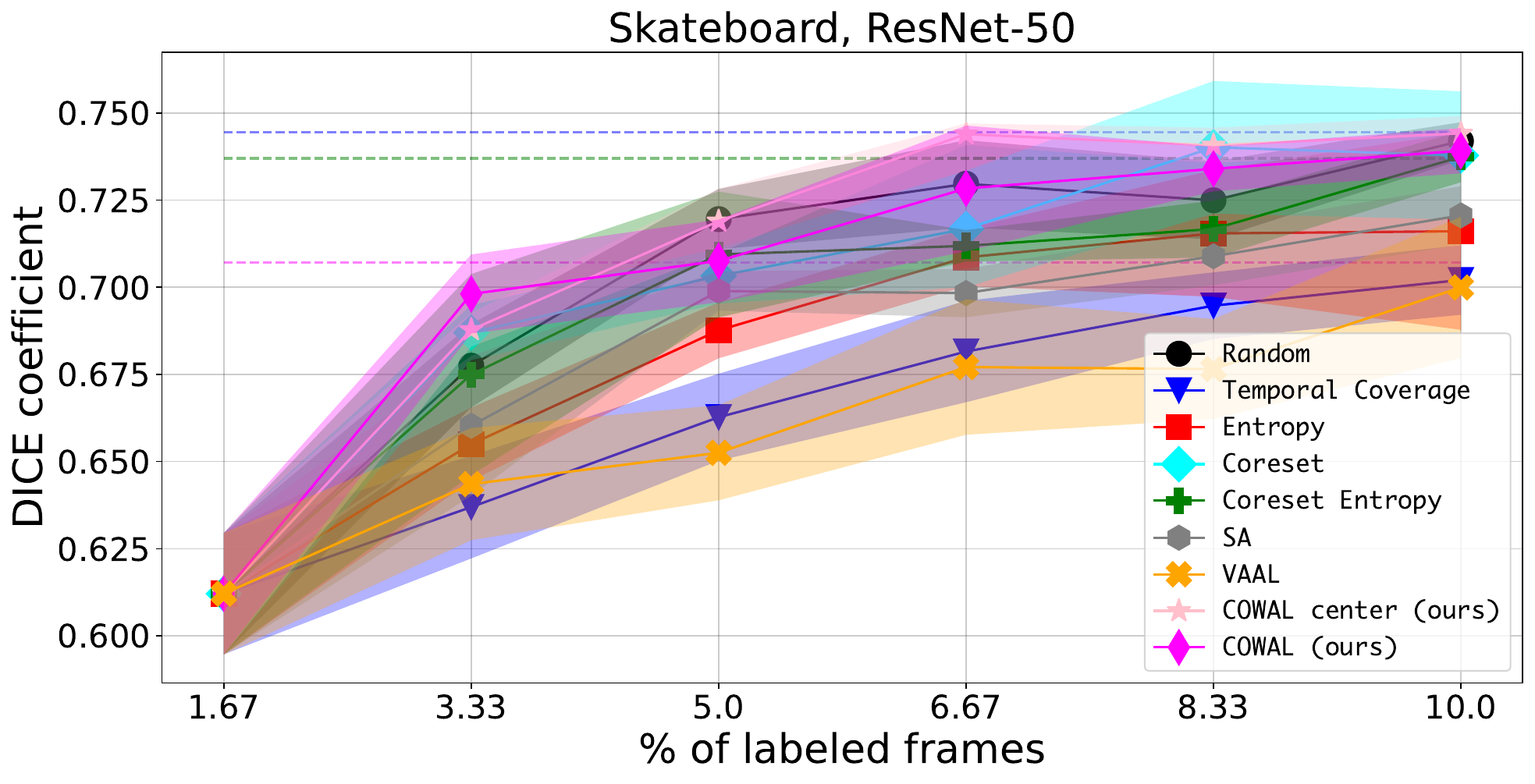}
    \includegraphics[width=0.49\linewidth]{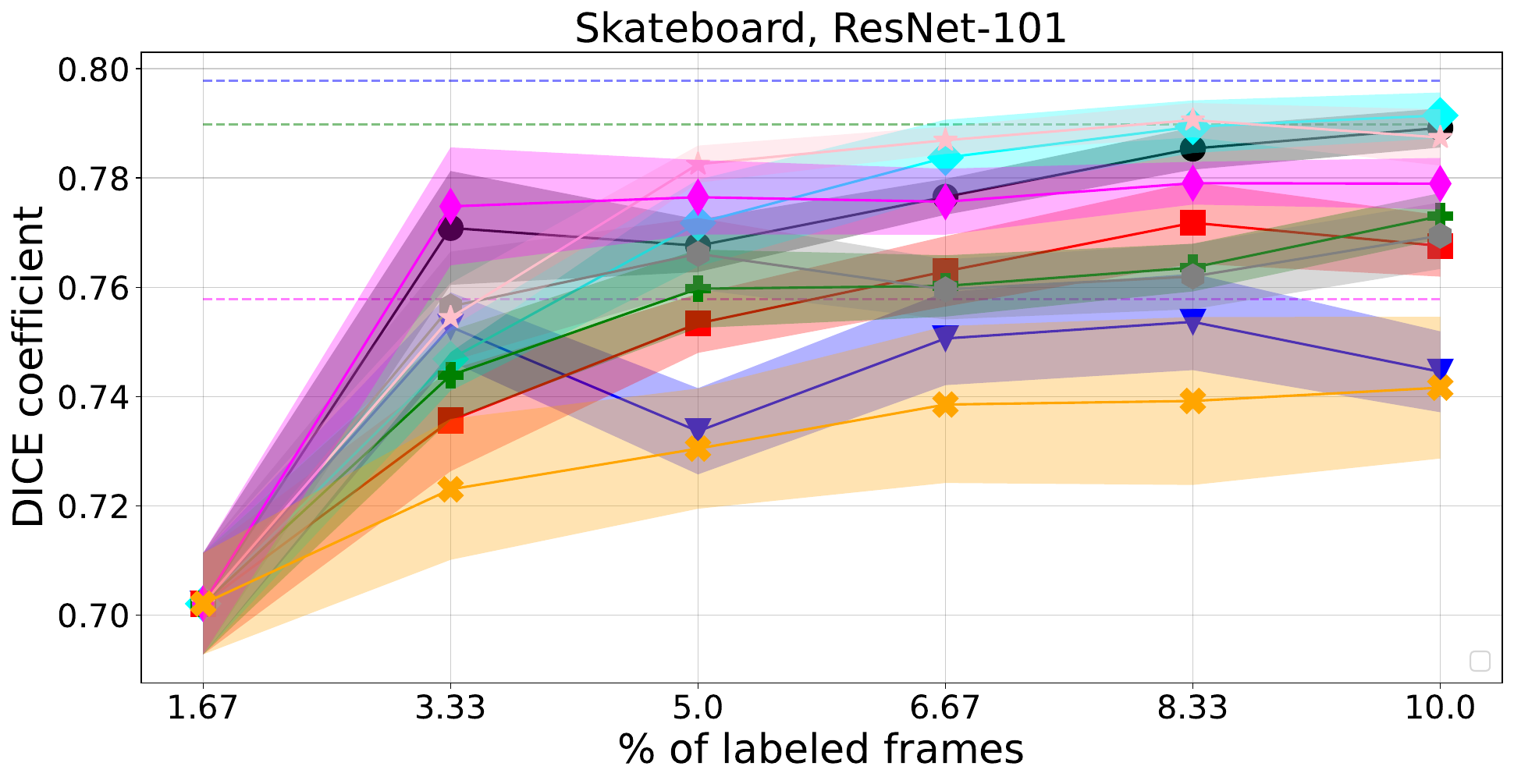}
    \medskip
    \includegraphics[width=0.49\linewidth]{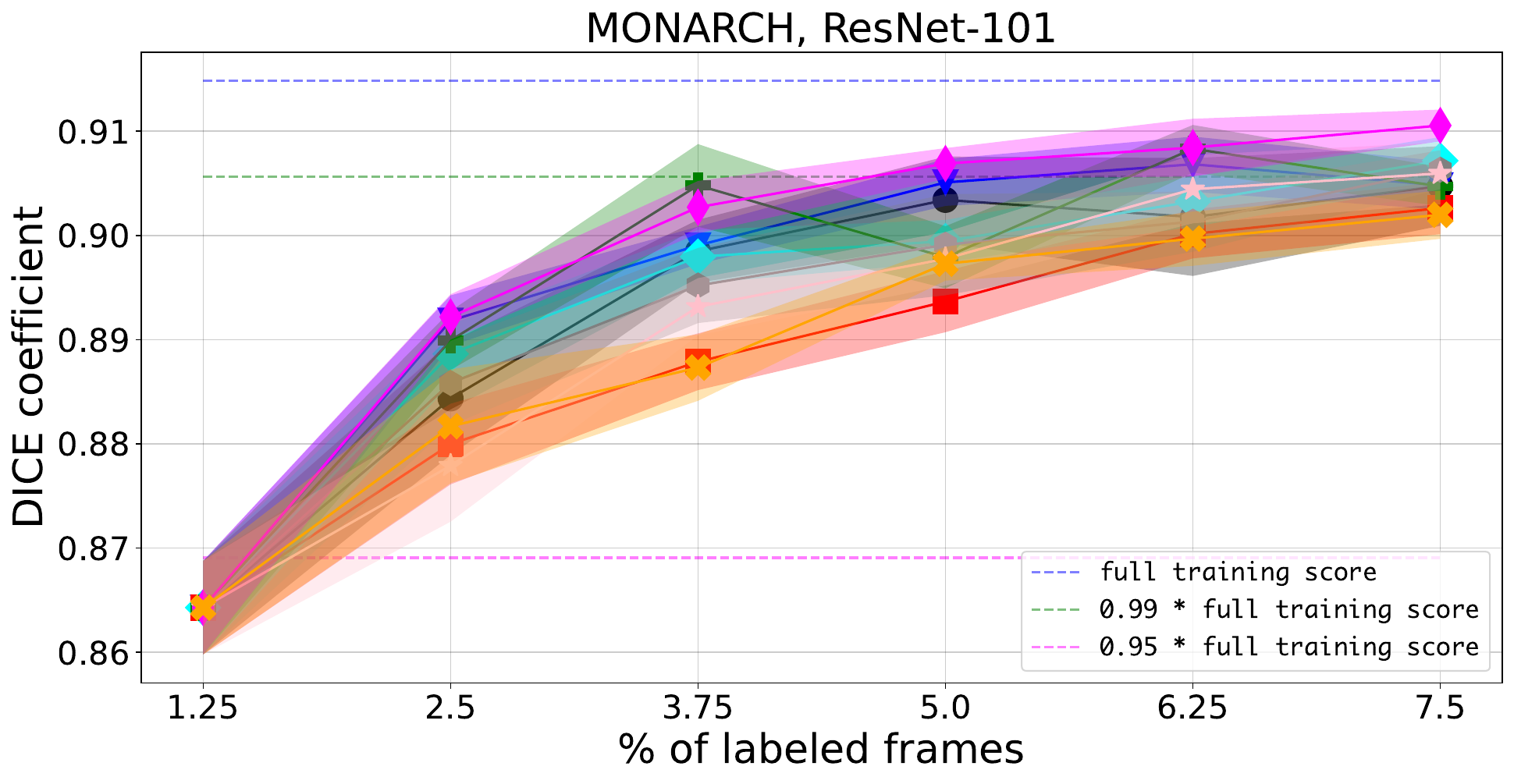}
    \includegraphics[width=0.49\linewidth]{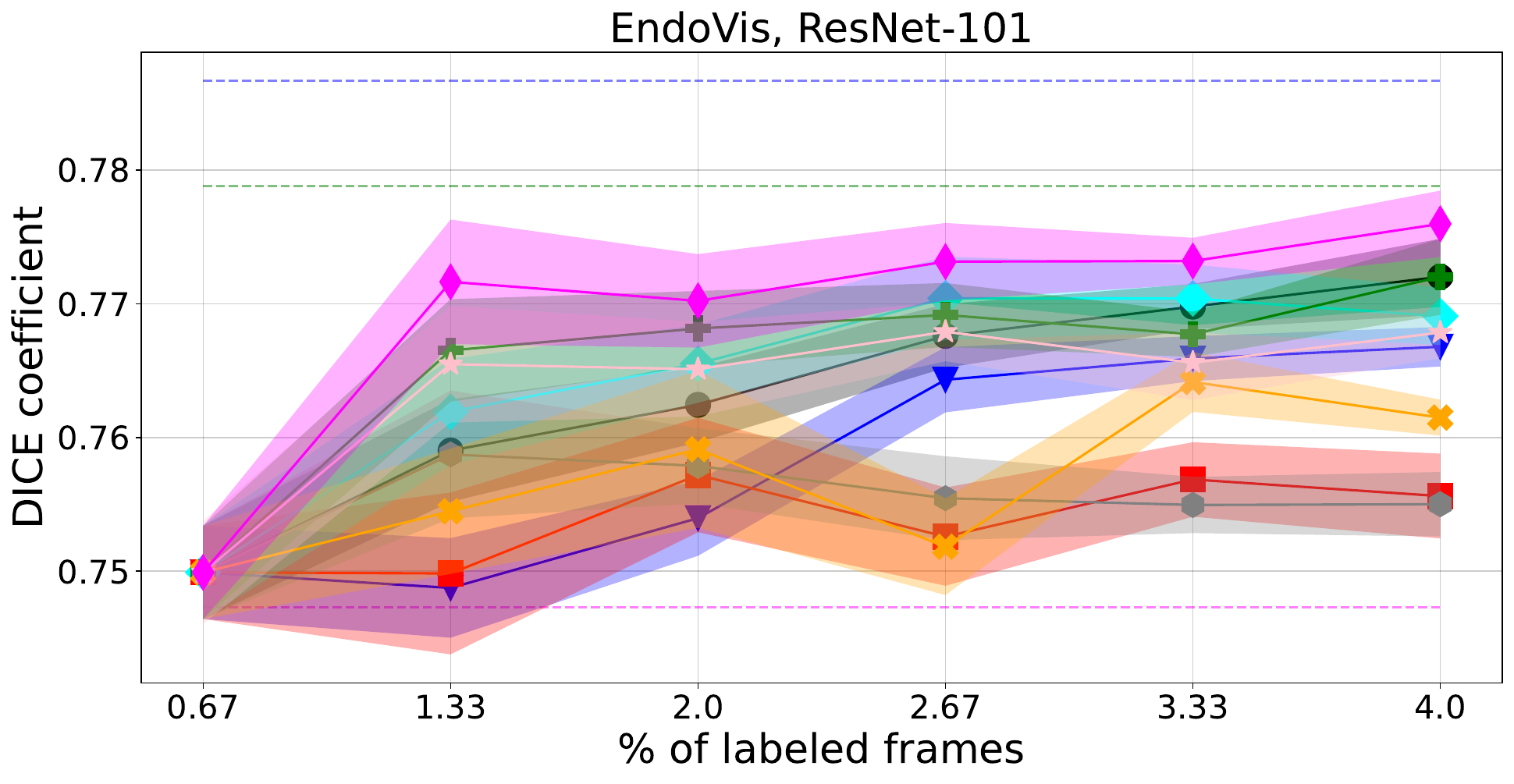}
    \medskip
    \includegraphics[width=0.49\linewidth]{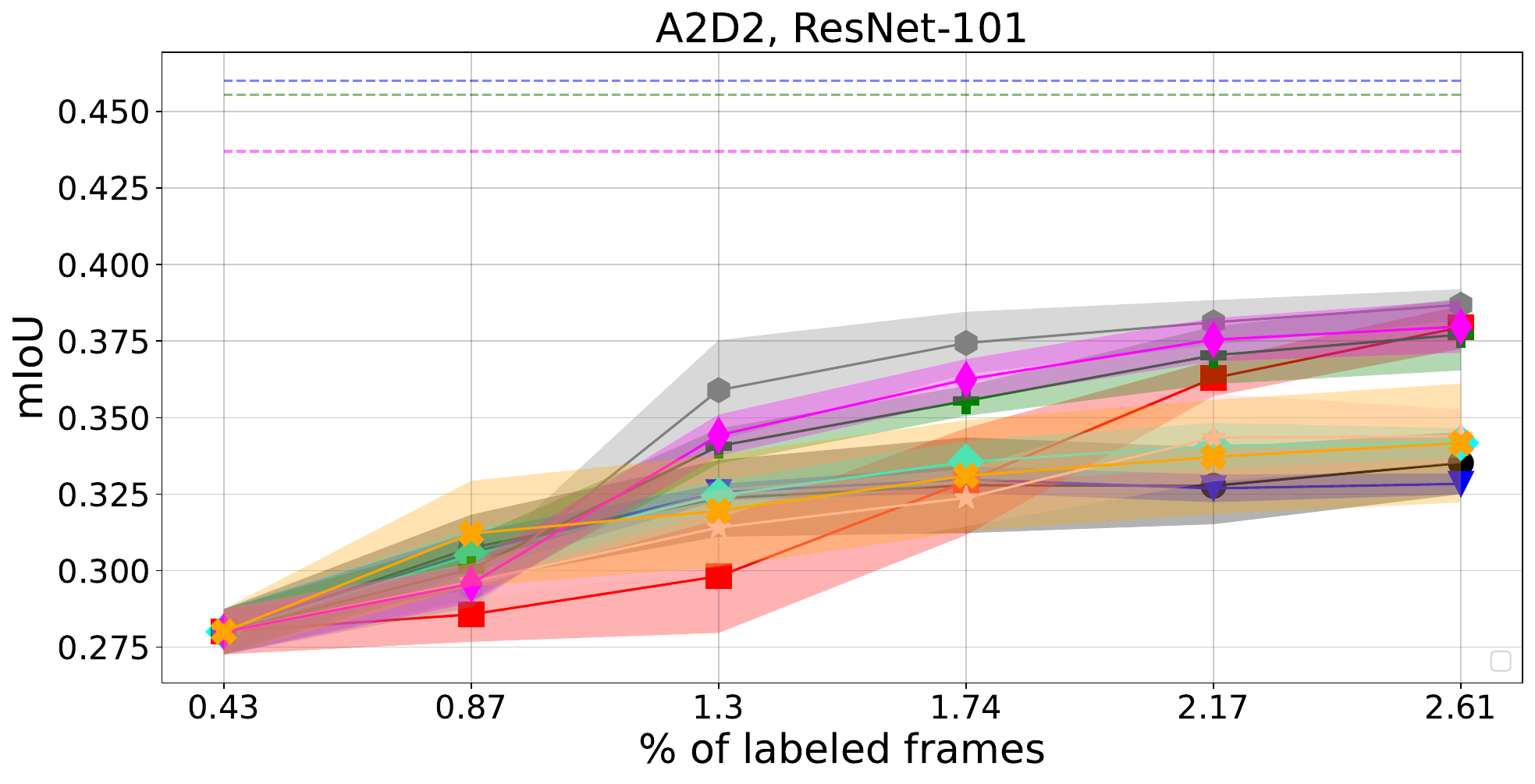}
    \includegraphics[width=0.49\linewidth]{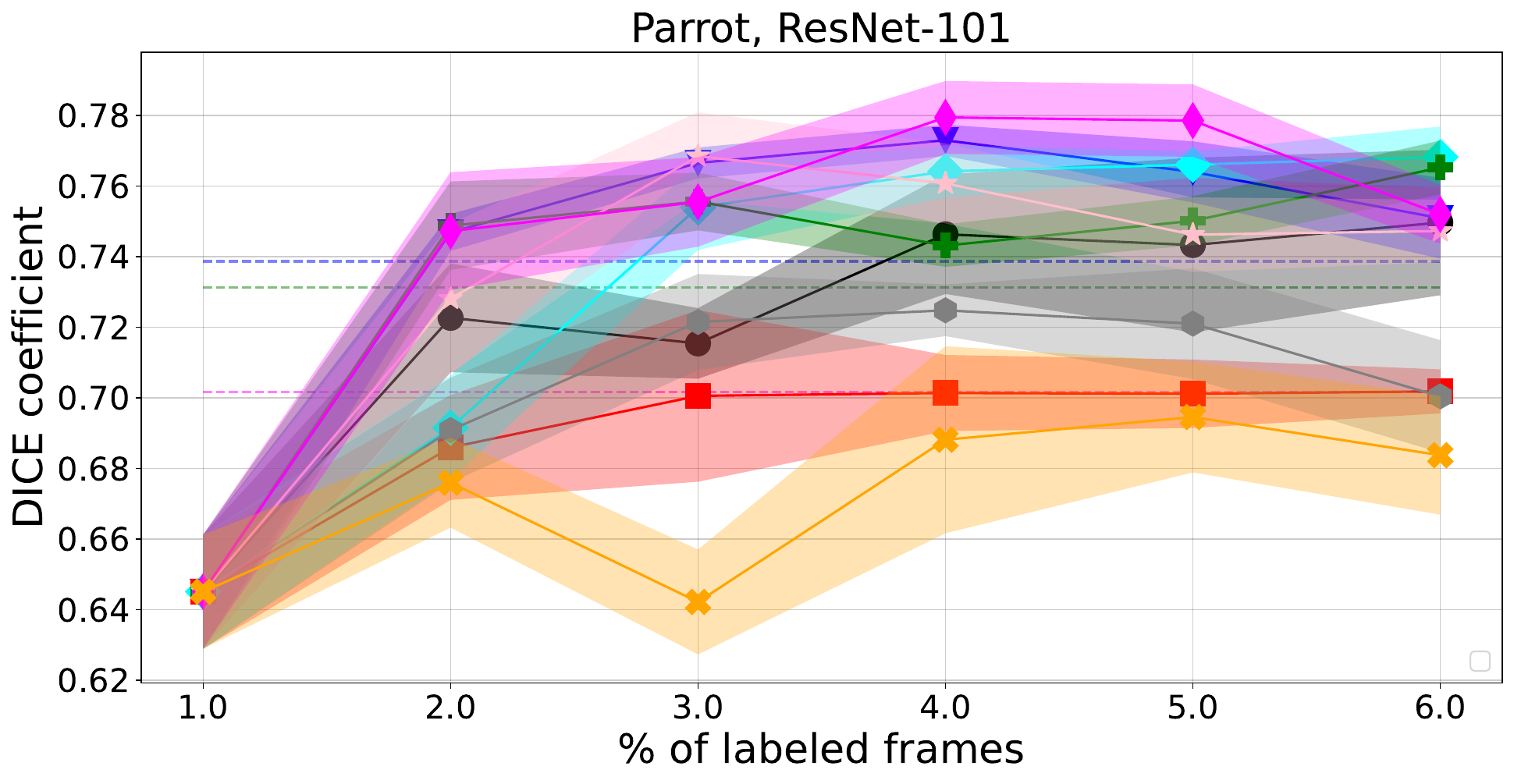}
   \caption{Appendix A. Comparison of different sampling strategies. Values are averaged over ten different training-validation splits, with error bars indicating one standard deviation.}
    \label{AL_extra_results}
\end{figure*}

Figure~\ref{AL_extra_results} shows the results for the skateboard dataset using the ResNet-50 backbone, along with the results for all datasets using the ResNet-101 backbone.

\begin{figure}
    \includegraphics[width=1\linewidth]{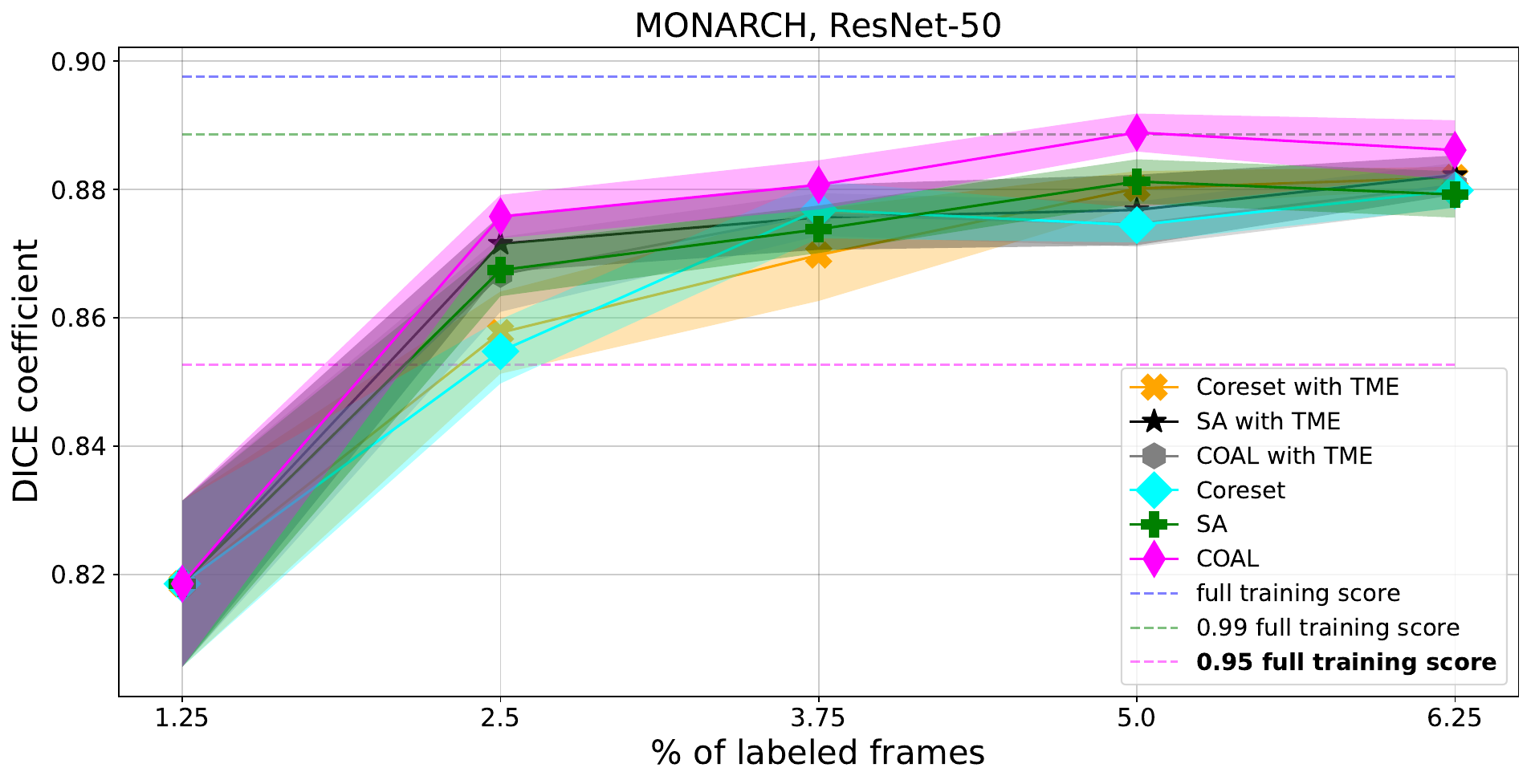}
    \includegraphics[width=1\linewidth]{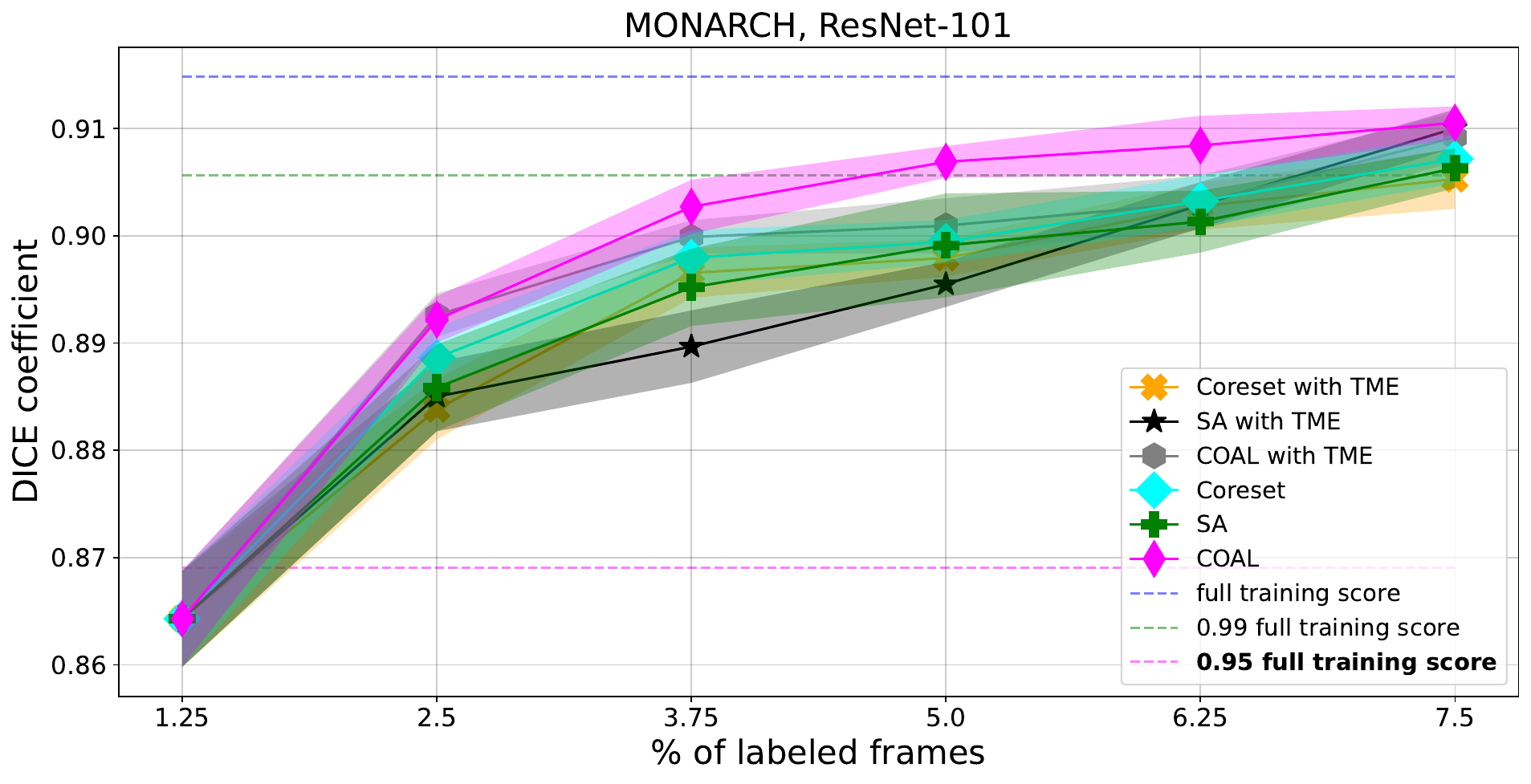}
    \includegraphics[width=1\linewidth]{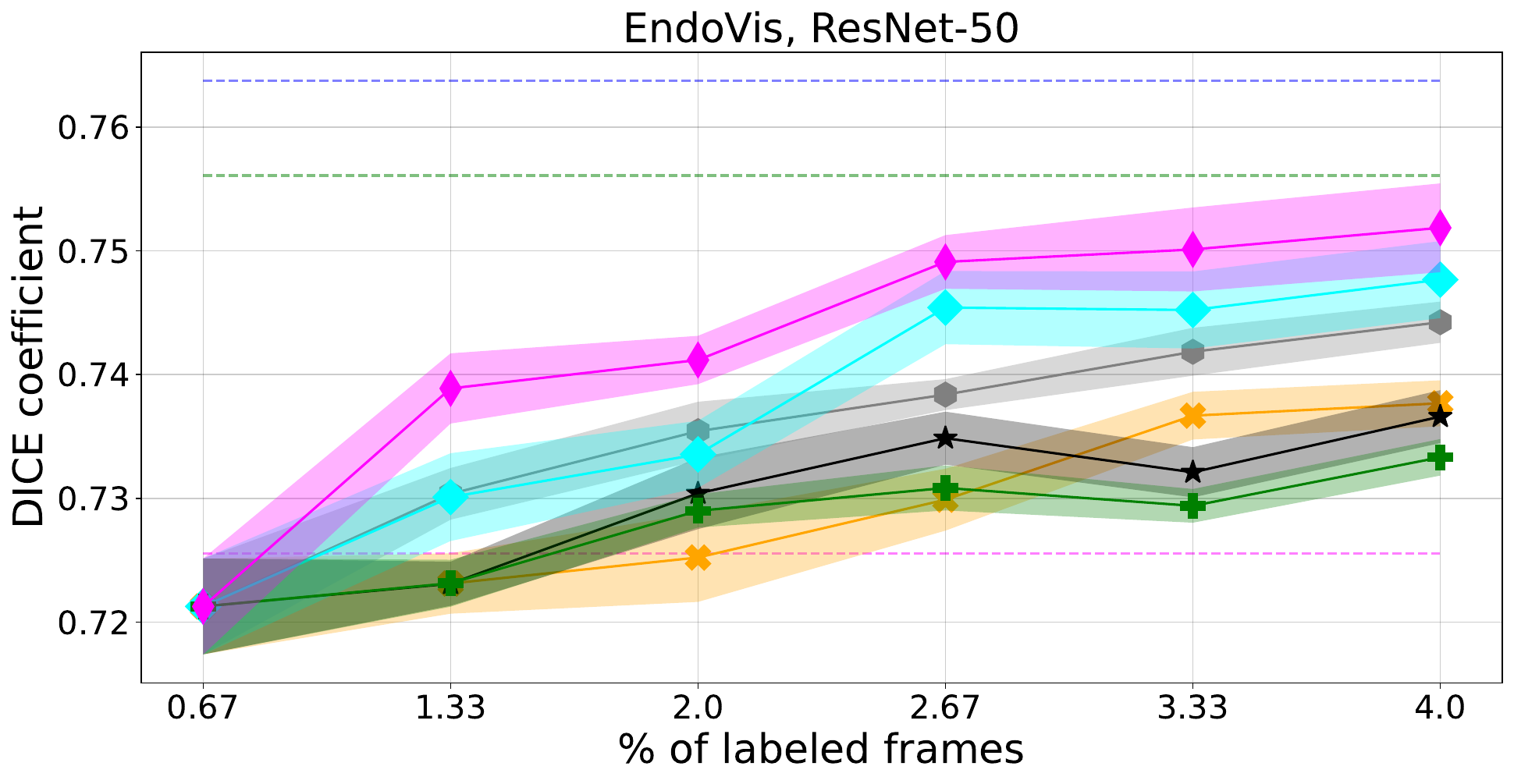}
    \includegraphics[width=1\linewidth]{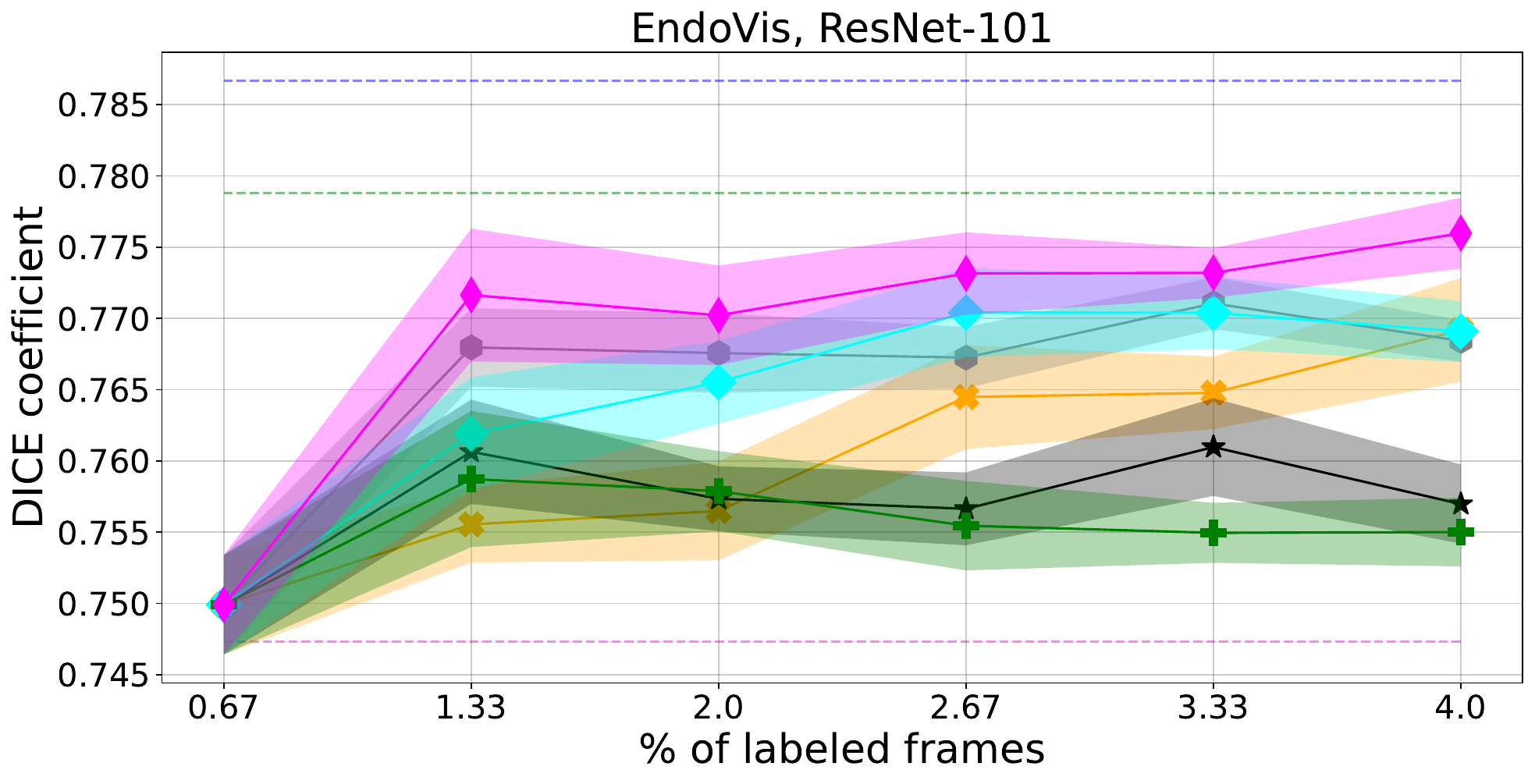}
   \caption{Comparison of \COALSamp{}, \CoreSamp{}, and \SASamp{} with two different embedding methods: Task Model Embedding (TME) or our contrastive embedding~$\phi$. Experiments are conducted on the MONARCH and EndoVis datasets with the ResNet-50 and ResNet-101 backbones. Evaluations are repeated for 10 different training-validation splits with error bars indicating one standard deviations.}
\label{embedding_extra_results}
\end{figure}

Figure~\ref{embedding_extra_results} shows additional comparisons between our contrastive embedding method and the standard task model embedding.

\section{Centroid matching algoritm}

Algorithm~\ref{centroid_matching} describes the procedure to find a matching between the embedding of labeled frames and the k-means centroids, as required by \COALSamp.

\section{Additional sampled frames}
\begin{figure*}[t]
    \includegraphics[width=1\linewidth]{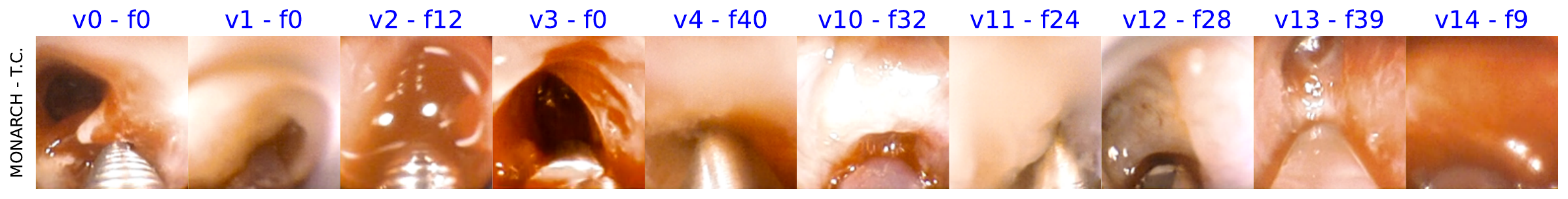}
    \includegraphics[width=1\linewidth]{figures/auris_AL_entropy_frames.pdf}
    \includegraphics[width=1\linewidth]{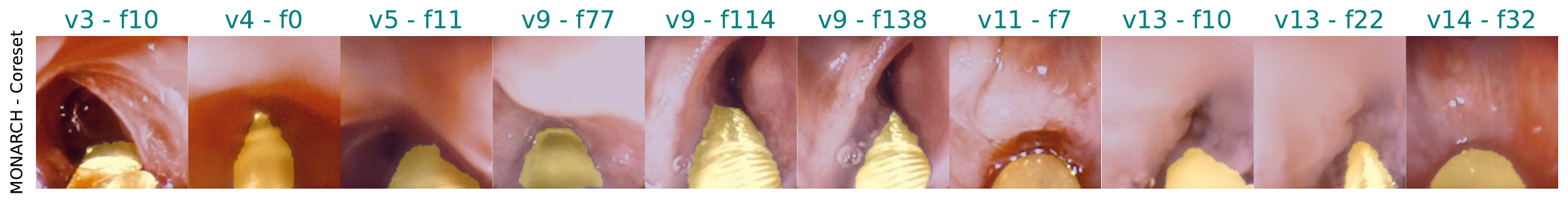}
    \includegraphics[width=1\linewidth]{figures/auris_AL_COWAL_entropy_simCLR_frames.pdf}
   \caption{
   Selected frames at the first iteration on the MONARCH dataset. Video and frame numbers are indicated on top of each image. 
   }
\label{fig:auris_frames}
\end{figure*}

\ref{fig:auris_frames} shows additional sampled frames by the \emph{Temporal Coverage}, \emph{Entropy}, \emph{CoreSet}, and \emph{COWAL} methods for the MONARCH dataset

\begin{figure*}[t]
    \includegraphics[width=1\linewidth]{figures/intuitive_AL_density_frames.pdf}
    \includegraphics[width=1\linewidth]{figures/intuitive_AL_entropy_frames.pdf}
    \includegraphics[width=1\linewidth]{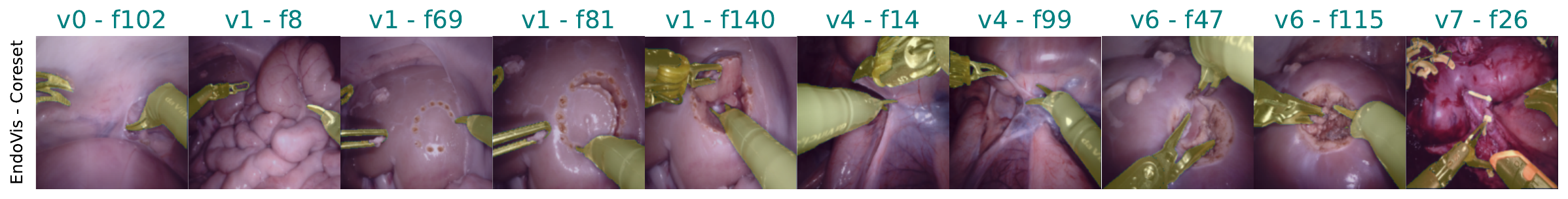}
    \includegraphics[width=1\linewidth]{figures/intuitive_AL_COWAL_entropy_simCLR_frames.pdf}
   \caption{
   Selected frames at the first iteration on the EndoVis dataset. Video and frame numbers are indicated on top of each image. 
   }
\label{fig:intuitive_frames}
\end{figure*}

\ref{fig:intuitive_frames} shows additional sampled frames by the \emph{Temporal Coverage}, \emph{Entropy}, \emph{CoreSet}, and \emph{COWAL} methods for the EndoVis dataset.

\begin{figure*}[t]
    \includegraphics[width=1\linewidth]{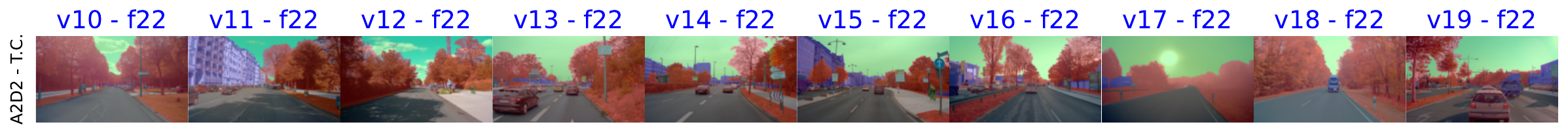}
    \includegraphics[width=1\linewidth]{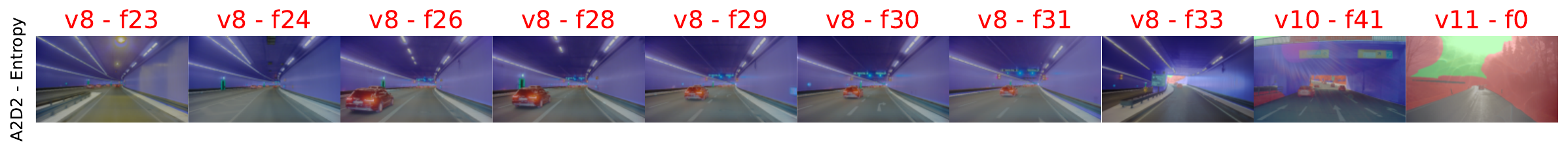}
    \includegraphics[width=1\linewidth]{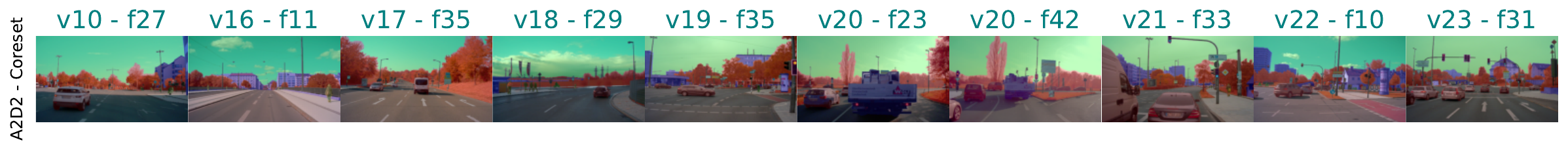}
    \includegraphics[width=1\linewidth]{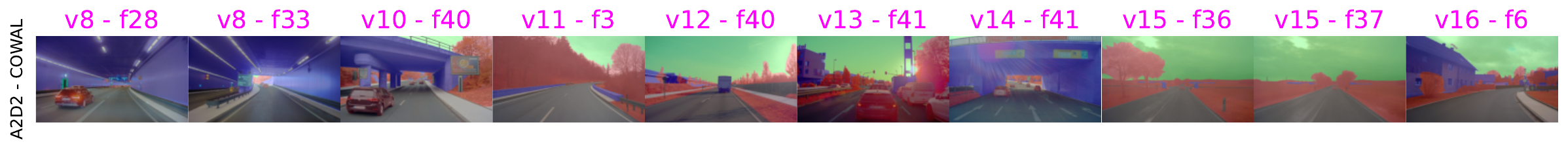}
    \includegraphics[width=1\linewidth]{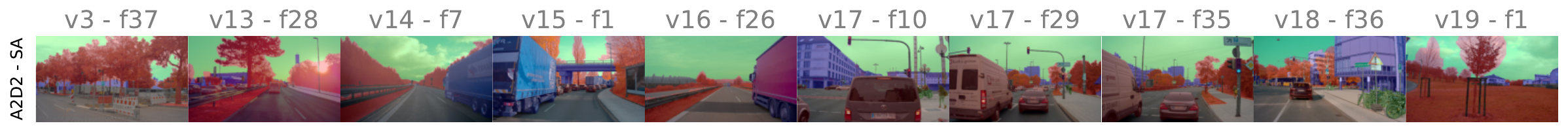}
   \caption{
   Selected frames at the first iteration on the A2D2 dataset. Video and frame numbers are indicated on top of each image. 
   }
\label{fig:a2d2_frames}
\end{figure*}

\ref{fig:a2d2_frames} shows additional sampled frames by the \emph{Temporal Coverage}, \emph{Entropy}, \emph{CoreSet}, \emph{Suggestive Annotation}, and \emph{COWAL} methods for the A2D2 dataset.

\begin{figure*}[t]
    \includegraphics[width=1\linewidth]{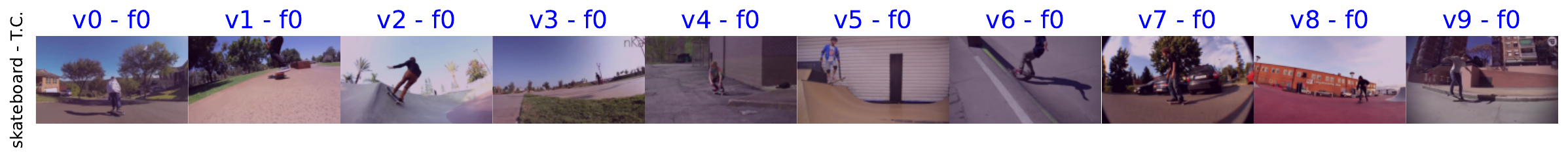}
    \includegraphics[width=1\linewidth]{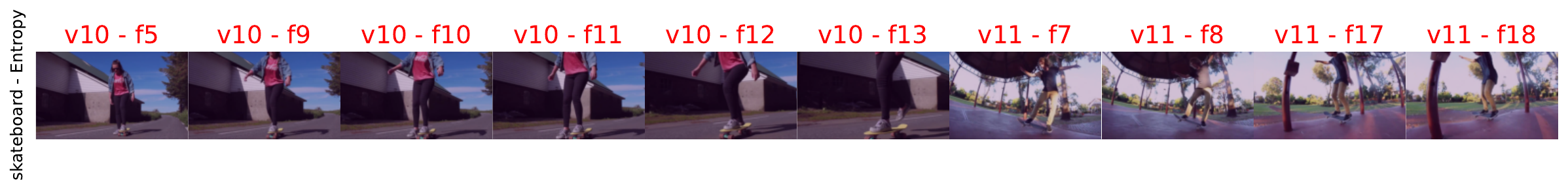}
    \includegraphics[width=1\linewidth]{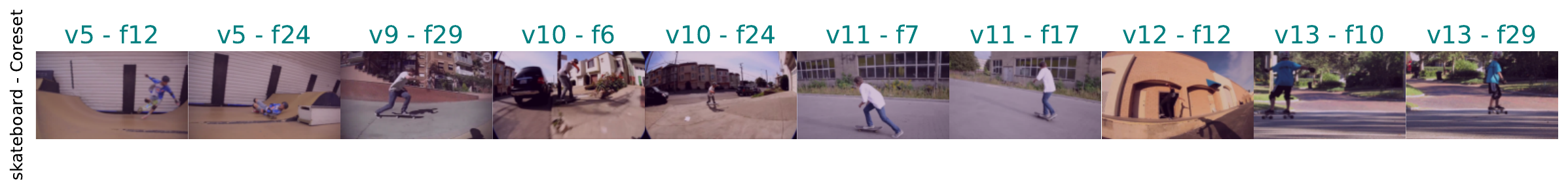}
    \includegraphics[width=1\linewidth]{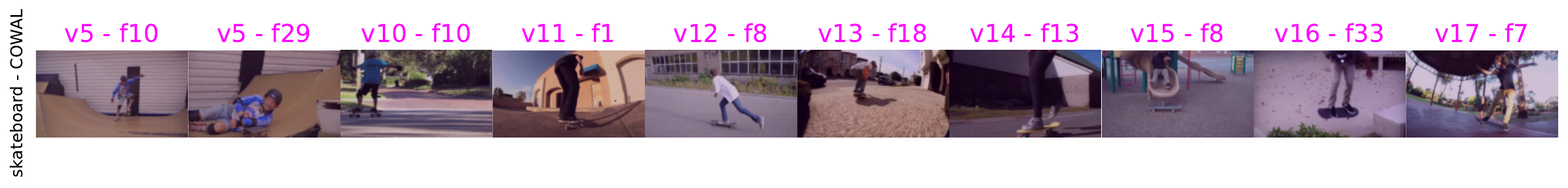}
   \caption{
   Selected frames at the first iteration on the Skateboard dataset. Video and frame numbers are indicated on top of each image. 
   }
\label{fig:skateboard_frames}
\end{figure*}

\ref{fig:skateboard_frames} shows additional sampled frames by the \emph{Temporal Coverage}, \emph{Entropy}, \emph{CoreSet}, and \emph{COWAL} methods for the Skateboard dataset.

\begin{figure*}[t]
    \includegraphics[width=1\linewidth]{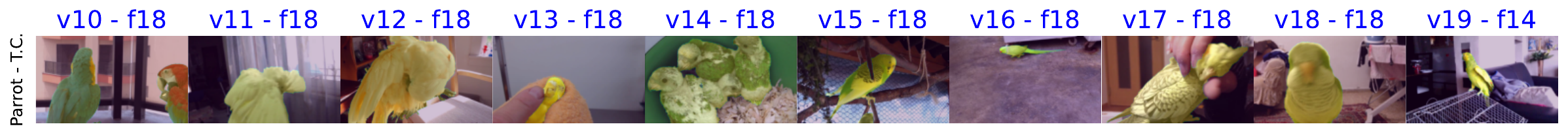}
    \includegraphics[width=1\linewidth]{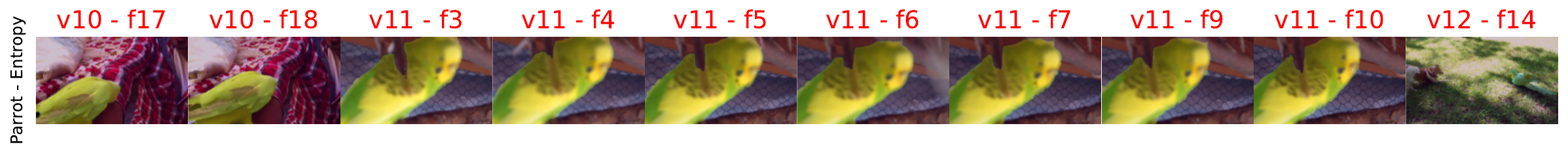}
    \includegraphics[width=1\linewidth]{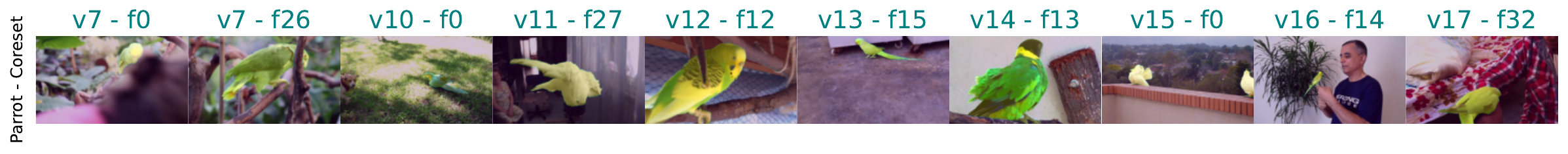}
    \includegraphics[width=1\linewidth]{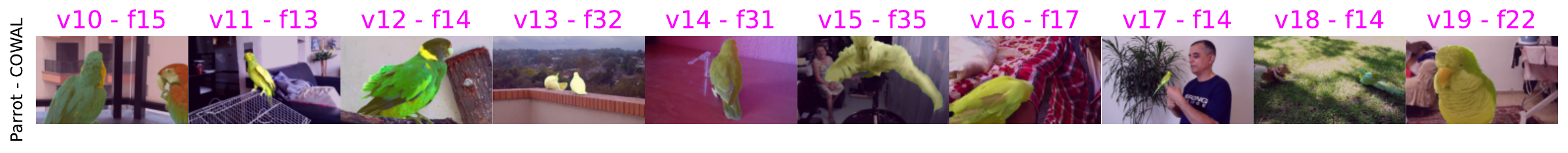}
   \caption{
   Selected frames at the first iteration on the Parrot dataset. Video and frame numbers are indicated on top of each image. 
   }
\label{fig:parrot_frames}
\end{figure*}

\ref{fig:parrot_frames} shows additional sampled frames by the \emph{Temporal Coverage}, \emph{Entropy}, \emph{CoreSet}, and \emph{COWAL} methods for the Parrot dataset.

\algnewcommand{\LineComment}[1]{\State \(\triangleright\) #1}

\begin{algorithm}
\caption{Centroid Matching}
\label{centroid_matching}
\begin{algorithmic}[1]
\Require Embedding of labeled frames $\{\a_i\}_{i=1}^{|\A|}$, K-means centroids $\{\k_j\}_{j=1}^K$ with $K=|\A|+Q$

\State $d[i, j]=\|\a_i - \k_j\|$
\Comment{Build matrix of pair-wise distances}
\State $\textrm{visit\_order} \leftarrow \textrm{argsort}_i \min_j d[i,j] $
\Comment{Labeled frames will be visited according to their distance to the closest centroid.}

\State $\mathcal{M}\leftarrow \emptyset$ \Comment{Set of matches}
\State $\textrm{assigned} \leftarrow \emptyset$ \Comment{Set of assigned centroids}
\LineComment{For each frame in the given order\ldots}
\For {$i'\in\textrm{visit\_order}$} 
\LineComment{Visit the centroids sorted by distance to $i'$}
\For {$j'\in \textrm{argsort}_j d[i', j]$} 
    \LineComment{Assign the first centroid not yet assigned}
    \If{$j'\notin \textrm{assigned}$}
        \State $\textrm{assigned} \leftarrow \textrm{assigned} \cup \{j'\}$
        \State $\mathcal{M} \leftarrow \mathcal{M} \cup \{(i', j')\}$
        \State \textbf{break}
    \EndIf
\EndFor
\EndFor

\LineComment{Return the set of matches}
\State \Return $\mathcal{M}$
\LineComment{A labeled frame~$i$ matches with centroid $m_i$ if $(i, m_i)\in\mathcal{M}$}
\end{algorithmic}
\end{algorithm}

\end{appendices}
%-------------------------------------------------------------------------
\clearpage
%%%%%%%%% REFERENCES
{\small
\bibliographystyle{ieee_fullname}
\bibliography{egbib}
}

\end{document}